\documentclass{IEEEoj}
\usepackage{cite}
\usepackage{amsmath,amssymb,amsfonts}
\usepackage{algorithmic}
\usepackage[ruled]{algorithm2e}
\usepackage{algorithmic}
\usepackage{graphicx,color}
\usepackage{float}
\usepackage{graphicx}
\usepackage{multirow,makecell}
\usepackage{paralist,url}
\usepackage{wrapfig}
\usepackage{tikz}
\usepackage{tikz-qtree}
\usetikzlibrary{trees,arrows,shapes,positioning,shadows}
\usetikzlibrary{positioning,shadows,trees,backgrounds,fit}
\usepackage{tikz}
 
\usepackage{amssymb}
\usepackage{pifont}
\newcommand{\xmark}{\ding{55}}%

\usepackage{verbatim}
\usepackage{glossaries}
\usepackage{subcaption,url}
 \usepackage{pgf}
\usepackage{amsmath}
\usepackage{multirow}
\usepackage{lscape}
\usepackage{textcomp}
\def\BibTeX{{\rm B\kern-.05em{\sc i\kern-.025em b}\kern-.08em
    T\kern-.1667em\lower.7ex\hbox{E}\kern-.125emX}}
\AtBeginDocument{\definecolor{ojcolor}{cmyk}{0.93,0.59,0.15,0.02}}
\def\OJlogo{\vspace{-14pt}\includegraphics[height=24pt,width=130pt]{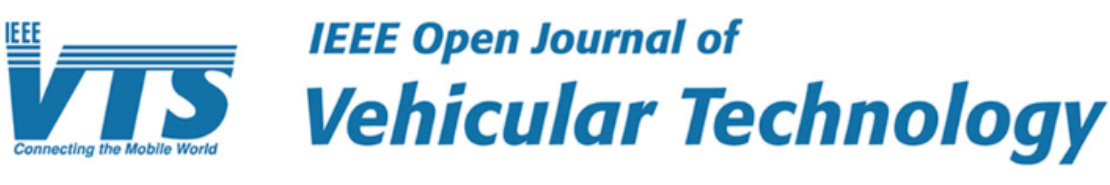}}
\begin{document}
\receiveddate{XX Month, XXXX}
\reviseddate{XX Month, XXXX}
\accepteddate{XX Month, XXXX}
\publisheddate{XX Month, XXXX}
\currentdate{XX Month, XXXX}
\doiinfo{OJIM.2022.1234567}

\title{Towards Reliable Participation in UAV-Enabled
Federated Edge Learning on Non-IID Data}

\author{Youssra Cheriguene$^{1,3}$, Wael~Jaafar$^{2}$, Senior Member, IEEE. Halim~Yanikomeroglu$^{3}$, Fellow, IEEE. and Chaker Abdelaziz Kerrache$^{1}$}
\affil{Laboratoire d’Informatique et de Mathématiques,
Université Amar Telidji de Laghouat, Laghouat, Algeria,}

\affil{Department of Software and Information Technology Engineering, École de Technologie Supérieure, Montreal, QC, Canada}
\affil{Non-Terrestrial Network (NTN) lab, Department of Systems and Computer Engineering, Carleton University, Ottawa, ON, Canada}

\corresp{CORRESPONDING AUTHOR: Youssra Cheriguene (e-mail: y.cheriguene@lagh-univ.dz).}

\markboth{Towards Reliable Participation in UAV-Enabled
Federated Edge Learning on Non-IID Data}{Youssra Cheriguene \textit{et al.}}

\begin{abstract}
Federated Learning (FL) is a decentralized machine learning (ML) technique that allows a number of participants to train an ML model collaboratively without having to share their private local datasets with others. 
When participants are unmanned aerial vehicles (UAVs), UAV-enabled FL would experience  heterogeneity due to the majorly skewed (non-independent and identically distributed -IID) collected data. In addition, UAVs may demonstrate unintentional misbehavior in which the latter may fail to send updates to the FL server due, for instance, to UAVs' disconnectivity from the FL system caused by high mobility, unavailability, or battery depletion. Such challenges may significantly affect the convergence of the FL model.
A recent way to tackle these challenges is client selection, based on customized criteria that consider UAV computing power and energy consumption. 
However, most existing client selection schemes neglected the participants' reliability. Indeed, FL can be targeted by
poisoning attacks, in which malicious UAVs upload poisonous local models to the FL server, by either providing targeted false predictions for specifically chosen inputs or by compromising the global model's accuracy through tampering with the local model.   Hence, we propose in this paper a novel client selection scheme that enhances convergence by prioritizing fast UAVs with high-reliability scores, while eliminating malicious UAVs from training.
Through experiments, we assess the effectiveness of our scheme in resisting different attack scenarios, in terms of convergence and achieved model accuracy. Finally, we demonstrate the performance superiority of the proposed approach compared to baseline methods.  
\end{abstract}




\maketitle

\begin{abstract}

Federated Learning (FL) is a decentralized machine learning (ML) technique that allows a number of participants to train an ML model collaboratively without having to share their private local datasets with others. 
When participants are unmanned aerial vehicles (UAVs), UAV-enabled FL would experience  heterogeneity due to the majorly skewed (non-independent and identically distributed -IID) collected data. In addition, UAVs may demonstrate unintentional misbehavior in which the latter may fail to send updates to the FL server due, for instance, to UAVs' disconnectivity from the FL system caused by high mobility, unavailability, or battery depletion. Such challenges may significantly affect the convergence of the FL model.
A recent way to tackle these challenges is client selection, based on customized criteria that consider UAV computing power and energy consumption. 
However, most existing client selection schemes neglected the participants' reliability. Indeed, FL can be targeted by
poisoning attacks, in which malicious UAVs upload poisonous local models to the FL server, by either providing targeted false predictions for specifically chosen inputs or by compromising the global model's accuracy through tampering with the local model.   Hence, we propose in this paper a novel client selection scheme that enhances convergence by prioritizing fast UAVs with high-reliability scores, while eliminating malicious UAVs from training.
Through experiments, we assess the effectiveness of our scheme in resisting different attack scenarios, in terms of convergence and achieved model accuracy. Finally, we demonstrate the performance superiority of the proposed approach compared to baseline methods.  
\end{abstract}
\section{Introduction}
\IEEEPARstart{M}{achine} achine Learning (ML)-assisted  techniques are garnering an increasing interest in various research disciplines, including unmanned aerial vehicle (UAV)-enabled wireless networks, as a result of the booming traffic data, the rising deployment of UAVs, and reduced costs \cite{9039589,Ghdiri2021,9825685}.
Specifically, UAVs are extensively employed for data collection and offloading \cite{9039589,Ghdiri2021}, which may be complex due to environment and user quality-of-service (QoS) requirements' unpredictable changes. Thus, the development of intelligent management schemes for related applications, e.g., based on ML, is needed. 
Conventional ML algorithms are cloud-centric, which makes them inappropriate for stringent QoS and UAV-enabled wireless networks. Indeed, relying on the cloud may expose sensitive information, such as the UAVs' locations and IDs, to attacks, besides increasing the UAVs' energy and bandwidth consumption needed for data transmissions.
To bypass this risk, federated edge learning (FEEL) has been recently introduced \cite{8970161,9694509}, where UAVs distributively train models on their local datasets without exposing their raw data through transmissions. 
However, UAV-based FEEL faces several challenges, such as statistical  heterogeneity and system heterogeneity. Indeed, due to the different locations of UAVs, collected data is significantly skewed, i.e., non-independent and identically distributed (non-IID). Moreover, the differences in UAVs' storage, computing, and power consumption, affect the latter's data collection and model training quality.  
Hence, system heterogeneity results in the appearance of FEEL stragglers, forcing the federated learning (FL) server to wait for the slowest UAVs prior to model aggregation, thus resulting in a slower FL convergence.  


In addition, UAV participants can demonstrate unintentional misbehavior during training due to poor connectivity or system faults. For instance, a UAV participant may fail to respond to the FL server due to its high mobility and/or low battery level, resulting in a small number of FL updates. In this case, the FL server would generate a biased global model and/or need to relaunch the FL round from the beginning.
Another unexplored issue is the potential intentional malicious behavior of UAV participants \cite{Yin2018ByzantineRobustDL,Pan2020JustiniansGR}. Indeed, such behavior may lead to denial-of-service (DoS) attacks on the FL server (untargeted attacks), or produce false predictions for FL targeted inputs (poisoning targeted attacks).
To tackle such problems, defense mechanisms have to rely on specific assumptions related to data distributions, e.g., data IIDness \cite{Pan2020JustiniansGR}, and the number of attackers \cite{NIPS2017_f4b9ec30,Yin2018ByzantineRobustDL}.  

The combination of the aforementioned FEEL risks would inevitably degrade the performances of FL, e.g., slow or even cancel model convergence, as well as increase training time, energy consumption, and bandwidth consumption. To the best of our knowledge, no prior work has jointly addressed the intentional and unintentional misbehavior of FL participants in UAV-enabled systems and its impact on FL performance.  
Consequently, we aim here to investigate the UAV participants' selection problem in heterogeneous data and system settings, in the presence of unreliable UAV participants categorized into one of the following profiles: Adversaries, stragglers, or dropouts. The classification of UAVs into these profiles is realized by analyzing their history and response behavior during training.
In the absence of malicious UAVs, i.e., adversaries, our proposed solution acts as an FL client selection scheme that prioritizes reliable UAVs. In contrast, in the presence of malicious UAVs, we target detecting and eliminating the latter in the early stages of FL training. The main contributions of this paper can be summarized as follows:

\begin{enumerate}
\item The diversity of locally available resources has an influence on overall learning performance, therefore, we propose a novel participant selection policy based on favoring highly reliable participants in order to eliminate training instability and reach the best FL accuracy performances.
\item In the presence of malicious participants performing untargeted or targeted attacks, we propose a line of defense that aims to identify and filter malicious updates/participants from training without requiring raw data exchanges.
\item Under the assumption of non-IID data distributions among UAV participants, we demonstrate through experiments that our proposed selection scheme efficiently mitigates unreliable participants and achieves superior performances, in terms of FL model accuracy, training time, dropout ratio, and minimal attack success rate, compared to baseline methods.

\end{enumerate}

The remainder of the paper is organized as follows.
Section \ref{sec: relatedwork} presents the related works. In Section \ref{sec:systemmodel}, the system model is introduced. Subsequently, the problem formulation and the proposed scheme are detailed in Section \ref{sec:problem-formulation}.
In Section \ref{sec: experiment}, experiments and performance evaluation are conducted. Finally,
Section \ref{sec: Conclusion} concludes the paper.

\section{Related Works}
\label{sec: relatedwork}
\subsection{Client Selection in FL}

Participants' resources can be extremely heterogeneous, e.g., different communication and computation capabilities, which may affect FL training \cite{8843900,9212434}, thus causing the appearance of stragglers. A straggler can be defined as an FL participant who spends a high time training its data and uploading its related local model to the aggregator or FL edge computing (EC) server. This slow training can be due to several factors such as the low computing power, the degraded communication link with the EC server, and the large size of the dataset. 
Indeed, the EC server should be receiving updates from participants in a reasonable time. 
On the other hand, a participant who leaves the FL training process before finishing its given tasks is referred to as a dropout. Dropping out from the FL system, i.e., quitting during the training or the aggregation phase in a given round, can significantly impact the FL performance. In fact, in such a situation, the EC server shall interrupt the current training round and relaunch it, thus leading to lost data and wasted energy consumption.
The dropping out event can be caused by intentional or unintentional quitting from the FL system. The latter can be due to the participant being  hijacked, or disconnected due to traveling away from the EC server.

To tackle the aforementioned issues, client selection has been developing in recent years.
In particular, for the stragglers' issue, client selection policies are set up to improve the overall training latency. For instance, the proposed client selection technique of \cite{273723}, called Oort, favored the participation of clients with the most useful data for both model accuracy enhancement and faster training. To mitigate the effect of unreliable participants, Oort removed clients that participated successfully in a given number of FL rounds. Nevertheless, this defense mechanism treated malicious and non-malicious participants equally, i.e., without any distinction. 


Subsequently, client clustering has been involved in the selection process to improve FL performances. For example,
authors of \cite{Chai2020TiFLAT} presented the TiFL algorithm, which divided participants into different tiers according to their training response latency. Then, participants belonging to the same tier would be selected at each training round. TiFL used an adaptive tier selection technique that adjusted clustering (or tiering) to the observed training performances over time.
In  \cite{10.1145/3511808.3557322}, the authors proposed FedRN, a robust FL method that exploits k-reliable neighboring participants with high data similarity to train the model only with clean samples over reliable participants. 

Due to the heterogeneity of computing and communication resources at FL participants, the authors of \cite{FedCS} tackled the stragglers' problem by proposing FedCS. FedCS is a greedy FL mechanism that favors incorporating more participants into its deadline-constrained training process, regardless of the datasets choice, and communication and computing capabilities.   

Given the complexity of FL systems, intelligent participant selection methods based on reinforcement learning (RL) have been proposed. For instance, a deep RL control method has been presented in \cite{2022} to accomplish the desired trade-off between local updating and global aggregation, in order to minimize the loss function while adhering to a resource budget limitation. 
In the context of FL for wireless networks, authors in \cite{9430906} examined the client scheduling problem to achieve low FL latency and rapid convergence, while taking into account the full or partial knowledge of channel state information (CSI) in communication links.
The scheduling problem has been solved using a multi-armed bandit (MAB) solution. Specifically, MAB was proposed to learn statistical data online without prior CSI knowledge and use it to build the scheduling strategy.
Similarly, federated reinforcement learning (FRL) was proposed in \cite{9945653} aiming to 
improve local driving models of connected and autonomous
vehicles (CAVs). Specifically, CAV selection for FRL has been investigated, while taking into account CAVs' reputation, quality of driving local models, and training time overhead.

To automate quality-aware FL participant selection, authors of \cite{9647925} proposed a novel scheme named AUCTION. The latter used reinforcement learning (RL) to progressively enhance the participants' selection policy through interaction with the FL platform after encoding the selection policy into an attention-based neural network. Although the developed method considered the quantity and quality of training data and cost of training, it ignored the effect of communication and computing latency, i.e., the stragglers' effect. Alternatively, 
the latter can be reduced using asynchronous FL where participants upload their local models asynchronously to the server \cite{https://doi.org/10.48550/arxiv.1911.02134,9093123,9356216}.



On the other hand, the dropout problem was investigated by only a few in the literature. For instance, \cite{9212434} proposed a multi-criteria-based client selection technique that combines for each client the computing power, storage, and energy consumption to forecast the latter's potential to carry out FL training rounds and handle the least number of discarded rounds caused by client dropouts.
When client dropouts are caused by highly dynamic environmental conditions, authors of \cite{BARBIERI2022100396} proposed decentralized
FL to reduce mandatory communications with the central FL server and rely instead on cooperative aggregation with nearby participants.
In \cite{9272671}, a mobile edge computing (MEC) framework is proposed for a multi-layer federated learning protocol, named HybridFL. This approach implemented model aggregation over two levels, namely edge and cloud levels, using several aggregation techniques. Moreover, it relied on a probabilistic method for participant selection at the edge layer to mitigate both stragglers and dropouts, while ignoring the state of participants (whose reliability is agnostic).
Through simulations, HybridFL is shown to shorten FL rounds' length and accelerate convergence.


In the aforementioned works, although participants may struggle to be performing well in FL, they have been considered benign or non-malicious, i.e., they do not misbehave intentionally against the FL process.

\subsection{Attacks on FL and Defense Mechanisms}

\begin{figure*}[t]
\centering
\includegraphics[trim={0cm 3.1cm 0cm 4.1cm},clip,width=0.999\linewidth]{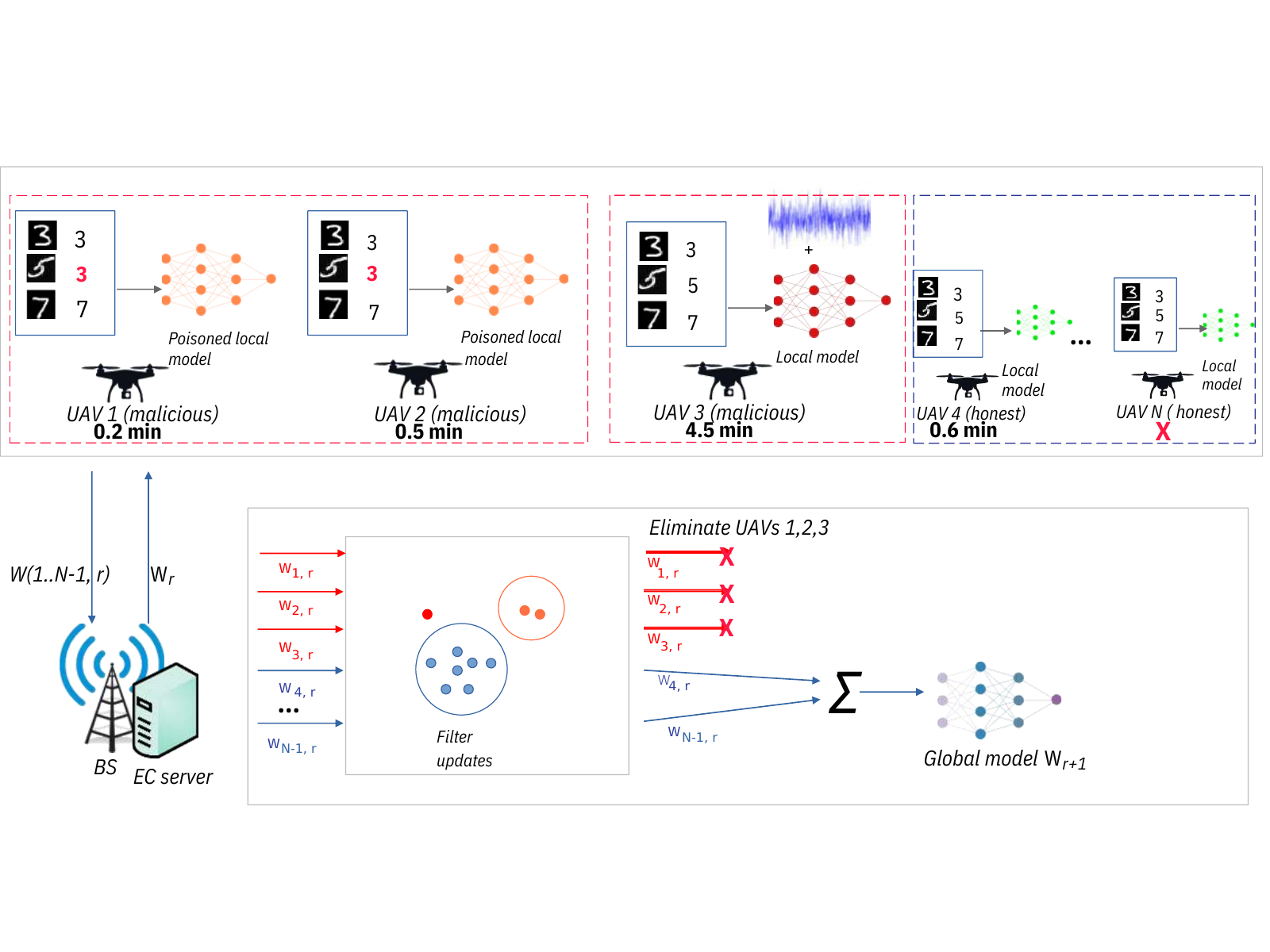}
\caption{System model.}
\label{fig:syst}
\end{figure*}

Despite FL's successful deployment in many fields, this ML technique is prone to attacks.
Indeed, given that participants have complete authority over the local training process, the latter becomes an attack surface through which an attacker, e.g., a malicious client, can access the learning process through the client system's vulnerabilities.
Subsequently, a poisoning (targeted) attack can be conducted, in which the ML model is altered during training, thus inducing the global model into producing erroneous outputs that correspond to the attacker's action on certain inputs, e.g., the attacker can conduct a data poisoning attack where it alters an image classifier to change an attacker-targeted label of local data samples with specific attributes. An example of a targeted attack is illustrated in Fig. \ref{fig:syst} for UAVs 1 and 2 \cite{pmlr-v108-bagdasaryan20a}.
Another type of attack is the untargeted attack, where the attacker aims to disrupt the training process by producing incorrect outputs on all inputs and thus slowing down FL convergence. We illustrate this case in Fig. \ref{fig:syst}  for UAV 3 \cite{Mallah2021UntargetedPA}.

\begin{figure}[t]
\centering
\includegraphics[trim={0cm 2.5cm 0cm 5cm},clip,width=0.999\linewidth]{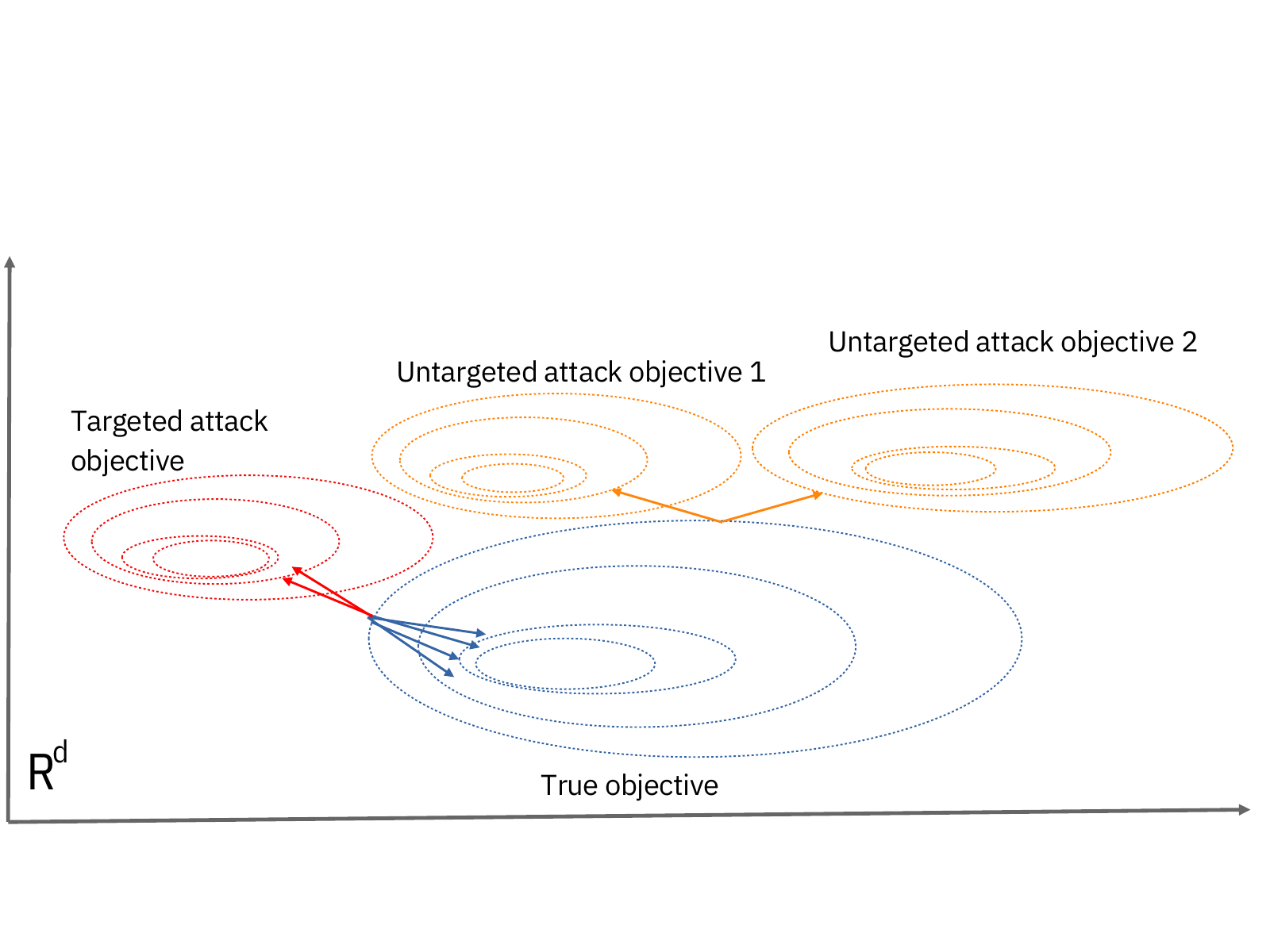}
\caption{Model attacks' objectives vs. true objective (Red and orange arrows point to the directions of the attacks' objectives, and blue arrows to the true objective).
}
\label{fig:objectives}
\end{figure}

For the sake of clarity, we detail the targeted and untargeted attack objectives in Fig. \ref{fig:objectives}. 
In the case of a targeted attack, attackers with a specific target or objective (red arrows) work together in each training round to advance towards the defined objective, by gradually shifting their malicious inputs. In an untargeted attack (orange arrows), attackers demonstrate different and even divergent behaviors, to achieve different local objectives.   

Since untargeted attacks have a direct and significant effect on the FL accuracy performance, they can be detected and mitigated through robust aggregation schemes performed at the server level \cite{10.5555/3294771.3294783,DBLP:journals/corr/abs-1803-01498}.
Similarly, a method called attestedFL was proposed in \cite{Mallah2021UntargetedPA} where each node's training history is monitored via a fine-grained assessment technique to detect malicious participants. 
However, the latter was proven inefficient in highly non-IID datasets.

For targeted attacks, authors of \cite{Awan2021CONTRADA} proposed a defense mechanism, called CONTRA, which uses a cosine-similarity-based metric and builds reputation by dynamically rewarding or punishing participants depending on their historical contributions to the FL global model. 
In \cite{Pan2020JustiniansGR}, authors presented Justinian's GAAvernor (GAA), a defense method that relies on RL to mitigate malicious attacks.
Specifically, GAA generated reward signals for policy learning using a quasi-validation set, i.e., a combination of a small dataset that exhibits a distribution similar to the actual sample distribution, and historical interactions data with the participants, as experience. However, this method was applied only for IID data samples, which is not realistic in the context of UAV-enabled systems.
In \cite{10.5555/3294771.3294783,10.1145/2991079.2991125}, model updates were clustered together, and the small-sized clusters were identified as malicious, thus eliminated from the FL process. Despite its simplicity, this approach would inevitably delete relevant data from non-malicious participants, due to clustering inaccuracy. 

Other methods relied on reputation design to avoid the negative impact of the participants' malicious behavior on FL performance. For instance, authors of \cite{9428537,9832523,8832210} proposed reputation-based participants' selection, where a reputation score is updated either according to the state and channel quality of each participant \cite{9428537}, or using reputation-aware clustering \cite{9832523}, or using an incentive mechanism to favor the frequent selection of highly reliable participants \cite{8832210}.

According to the above discussion, there is a need to mitigate unreliable participants in FL heterogeneous systems. Table \ref{Tab0} below summarizes the key differences between related works and our contribution. To the best of our knowledge, existing works are not effective in defending against unreliable participants demonstrating different behaviors. 
Furthermore, most works can not be applied directly to UAV-based systems due to the latter's heterogeneity. 

\begin{table*}[!h]
\centering
\caption{Comparison of related works.}
\label{Tab0}
\begin{tabular}{|c|c|c|c|c|c|c|}
\hline

\multirow{2}{*}{\textbf{Reference}} & 
\multirow{2}{*}{\makecell{\textbf{Dataset} \\ \textbf{Setting}}} 
& \multicolumn{4}{c|}{\textbf{Features}} & 
\multirow{2}{*}{\textbf{Defense  strategy}} 
\\ 
 \cline{3-6} 
 {} & {} & \makecell{\textbf{Dropouts}} & \makecell{\textbf{Stragglers}} &  \makecell{\textbf{Client}\\ \textbf{Selection}} & \makecell{\textbf{Attack} \\ \textbf{type}} & {}\\ \hline

\makecell{\cite{273723}}  & \makecell{Non-IID}   & \xmark &  \ding{51} &  \ding{51} & \xmark  & \makecell[l]{Pre-defined maximum number of participation rounds \\per client.}\\
 \hline

\makecell{ \cite{Chai2020TiFLAT}, \cite{2022},
\cite{9430906}, \cite{FedCS}, \\ \cite{9945653}}  & \makecell{Non-IID} & \xmark  &  \ding{51} &  \ding{51} & \xmark & \xmark \\
 \hline
   \makecell{\cite{9212434}, \cite{BARBIERI2022100396}, \cite{9272671}}  & \makecell{Non-IID}  & \ding{51} &  \ding{51} &  \ding{51} & \xmark & \xmark \\
 \hline
  \makecell{\cite{10.1145/3511808.3557322}}  & \makecell{Non-IID}  & \xmark &  \xmark & \ding{51} & \xmark & \xmark \\
 \hline

 \makecell{\cite{9647925}}  & \makecell{Non-IID}  & \xmark &  \xmark  & \ding{51} & \xmark & \xmark\\
 \hline

 \makecell{\cite{https://doi.org/10.48550/arxiv.1911.02134,9093123,9356216}}  & \makecell{Non-IID}  & \xmark &  \ding{51}  & \ding{51} & \xmark & \xmark\\
 \hline
 
 \makecell{ \cite{10.5555/3294771.3294783}, \cite{DBLP:journals/corr/abs-1803-01498}}  & \makecell{IID}  & \xmark &  \xmark &  Random & Untargeted  & \makecell[l]{Robust aggregation based on median statistics of received\\ gradients.}\\
 \hline
 \makecell{ \cite{NIPS2017_f4b9ec30}}  & \makecell{IID}  &  \xmark &  \xmark &  Random & Untargeted & \makecell[l]{The gradient with the closest distance 
to its neighboring\\ gradients is selected as the global gradient.} \\
 \hline
  \makecell{\cite{Mallah2021UntargetedPA}
}  & \makecell{IID}  &  \xmark &  \xmark &  Random & Untargeted & \makecell[l]{Use of cosine similarity between models sent by the
same\\ participant + use of quasi-validation dataset prior \\to aggregation.} \\
 \hline
   \makecell{ \cite{Pan2020JustiniansGR}
}  & \makecell{IID}  &  \xmark &  \xmark &  Random & Targeted & \makecell[l]{RL technique that uses participants' interactions as \\experience and a quasi-validation dataset for reward signals.} \\
 \hline
  \makecell{\cite{Awan2021CONTRADA}}  & \makecell{Non-IID}  & \xmark &  \xmark &  \ding{51} & Targeted & \makecell[l]{Using a reputation metric, the global model rewards or \\punishes particular participants depending on their \\past contributions.} \\
 \hline

  \makecell{ \cite{9428537} }  & \makecell{Non-IID}  & \xmark &  \xmark &  \ding{51} & Targeted & \makecell[l]{Use of a reputation model based on beta
distribution function\\ to determine the credibility of local participants.}\\
 \hline
  \makecell{ \cite{9832523} }  & \makecell{Non-IID}  & \xmark &  \xmark & \ding{51} & Targeted & \makecell[l]{Use of a reputation-aware FL participant selection  method \\ based on stochastic integer programming.}\\
 \hline

  \makecell{ \cite{8832210} }  & \makecell{Non-IID}  & \xmark &  \xmark & \ding{51} & Targeted & \makecell[l]{The combination of a multi-weight subjective logic model \\with a  reputation-based participant selection approach \\for reliable FL.}\\
 \hline

  \makecell{\cite{9622905}
}  & \makecell{IID}  & \xmark &  \xmark & Random & Targeted & \makecell[l]{Use of
dimensionality reduction techniques + K-means\\ clustering of received updates.}\\
 \hline
 
  \makecell{\textbf{This}\\ \textbf{work}}  & \makecell{Non-IID} & \ding{51} &  \ding{51} &  \ding{51} & \makecell{Targeted \\+ untargeted} & \makecell[l]{IQR-based straggler filtering + reliability-based dropouts \\filtering + DBSCAN clustering of cosine similarity distances\\ between received updates for malicious UAVs filtering.}\\
 \hline
\end{tabular}
\end{table*}



\section{System Model}
\label{sec:systemmodel}
The considered system model is depicted in 
Fig. \ref{fig:syst}. It is composed of a base station (BS) that is paired to an edge computing (EC) server to satisfy ground users' and $N_{U}$ UAVs' FL tasks. Ground users and $N_{U}$ UAVs are deployed in the BS's 3D coverage area. 
The UAVs act as edge nodes with lower capabilities than the BS, but sufficiently high to perform local ML training on their local datasets, denoted as $DS_{i}$, $\forall i=\{1,\ldots, N_U \}$\footnote{We assume here that UAVs have sufficient power to execute ML while flying. This can be justified by the advancements in powerful miniaturized computing devices with low-power consumption, e.g., tiny ML \cite{parmar2023memoryoriented,ray2023tiny}.}. For the sake of simplicity, we assume here that the FL process is set up among the BS as the aggregator and only the UAVs as participants, where the model's local update parameters, denoted by weight vector $w_{i,r}$ for UAV $i$ in FL round $r$, are uploaded from the UAVs to the BS.    
Moreover, we assume that the EC server has four engines: A participant filtering engine, a participant selection engine, a security engine, and an aggregation engine.
The participant filtering engine identifies and eliminates the stragglers in the participants' pool.
The participants' selection engine chooses the UAVs that will train and upload their model parameters in each FL round. 
The security engine is used to eliminate unreliable participants, while the aggregation engine performs model aggregation on the local updates of selected UAVs into a global model, as shown in Fig. \ref{fig:syst}. Furthermore, for ease of reading, notations are described in Table \ref{notations}.
\begin{table}[!h]
\caption{Notations.}
    
\begin{tabular}{c l } 

 \textbf{Notation} & \textbf{Description} \\ 
$N_U$  & Number of UAVs \\ 
$r_0$  & Number of global rounds \\ 
$w_{i,r}$  & Weight vector of client $i$ in round $r$\\ 
$w_{r}$  & Aggregated global model in round $r$\\ 
$M_{r,i}$ & Number of data samples of client $i$ \\
$t_{i,r}$  & Local training time of client $i$ in round $r$\\ 
$t_{i,BS}$  & Upload time of local updates of client $i$ \\ 
$t_{BS,i}$  & Download time of global model to client $i$ \\ 
$t_{r}$  & Total time for round $r$\\ 
$t_{ag}$ & Aggregation time on weight updates \\
$t_{r}^i$ & Training and uploading time of client $i$ in round $r$ \\
$T_e$ & Number of local iterations \\
$\kappa$ & Number of CPU-cycles per data sample \\
$\gamma_i$ & CPU frequency of client $i$ \\
$d_{i}$ & 3D distance between UAV $i$ and the BS \\
$\beta_0$ & The reference channel gain \\
$h_{i}$ & Communication link between UAV $i$ and the BS \\
$R_{i,BS}$ & Uplink data rates between UAV $i$ and the BS \\
$R_{BS,i}$ & Downlink data rates between UAV $i$ and the BS \\
$B_i$ & Bandwidth reserved to UAV $i$ by the BS\\
$P_i$ & Transmission power of UAV $i$ \\
$P_{BS}$ & Transmission power of the BS \\
$V$ & The number of NN model parameters \\ 
$z$ & The size of a model parameter \\
$\sigma^2$ & The unitary power of the additive white Gaussian noise \\
$\alpha_{i}$ & Path-loss component \\
$A_{i,r}$  & Accuracy of local model of client $i$ in round $r$\\ 
$A_{g}$  & Global accuracy after $r_0$ rounds of training \\ 
$\Lambda_{i,r}$  & Reliability score of client $i$ in round $r$\\ 
$\rho$ & Min threshold value for the reliability score \\ 
$\rho_0$ & Max threshold of the reliability score\\
$W_{r}$  & Received weight updates in round $r$ \\ 
$\Gamma_{i,r}$ & Score of client $i$ in round $r$ \\
$\kappa_{i,j}$ & Cosine similarity between weight updates $i$ and $j$ \\
$\hat{w}_{i,r}$ &  Weight update drift of client $i$ in round $r$ \\
$P_r$ &  Participating UAVs in round r \\
$\mathcal{P}_r^h$ & Set of honest UAVs in round $r$ \\
$\mathcal{P}_r^s$  & Set of straggler UAVs in round $r$ \\
$\mathcal{P}_r^d$ & Set of dropout UAVs in round $r$ \\
$\mathcal{P}_r^m$ & Set of malicious UAVs in round $r$ \\
$\mathcal{\tilde{P}}_r$ &  Participating UAVs in aggregation in round $r$ \\
$\nu$ & Upper bound scaling parameter \\
$f$ & Attackers ratio in the UAV network \\
$\chi_d$ & Dropout ratio in the UAV network \\
$f_{r}$ & Minimum UAV participants fraction \\
$\tau_{a}$ & Average FL round time  \\

\end{tabular}
\label{notations}
\end{table}


\subsection{Federated Learning Model}
\label{sec: FL}
Initially, the EC server creates a generic model for a specific FL task, then selects a set of participants with whom to communicate model parameters. The minimum number of participants selected is  ${{P}}_r=N_{U} \times f_{r}$, where $f_{r} \in [0,1]$ is the fraction of participants involved in round $r$. Once selected, UAV client $i$ downloads the most recent global model $w_{r-1}$ from the EC server for synchronization. Then, it executes multiple iterations of stochastic gradient descent (SGD) on mini-batches $M_{i,r}$ from its dataset $DS_i$, to update its model weight $w_{i,r}$, such that 
\begin{equation}
\label{eq:SGD}
  w_{i,r} = SGD(w_{r-1},M_{i,r}) - w_{r-1},\; \forall i=1, \ldots,P_r.
\end{equation}
Periodically, the resulting weight update is uploaded to the EC server.
Subsequently, the EC server aggregates the collected local models' parameters to produce the global model $w_{r}$ as follows: 
\begin{equation}
\label{eq:globalmodel} 
{w}_{r}= w_{r-1}+\sum_{i=1}^{P_r} \frac{|M_{i,r}|}{|M_r|} w_{i,r},
\end{equation}
where $M_r=\cup_{i=1}^{P_r} M_{i,r}$ denotes the number of data samples from $P_r$
participants in round $r$.
If an insufficient number of model updates is received, i.e., a high dropout ratio, the round is canceled by the EC server.

\subsection{FL Latency Model}
Based on the hardware
configuration information and size of the local training dataset of the $i^{th}$ UAV,
the EC server can estimate the local training time as \cite{9052206}
\begin{equation}
\label{eq:comp}
    t_{i,r}=T_e \frac{\kappa  |M_{i,r}|}{\gamma_i}, \; \forall i =1,\ldots,P_r, 
\end{equation}
where $\kappa$ is the number of CPU-cycles needed to process one data sample, $T_e$ is the number of local iterations,
and $\gamma_i$ is the available CPU-frequency of UAV client $i$.

 At the end of $T_e$ local training iterations, the neural networks' (NN) updated parameters are transmitted through the wireless channel to the EC server, via the BS.

{Assuming that communications between the BS and UAVs follow the free-space air-to-ground path loss model, that BS and UAVs are equipped with a single-antenna, and that simultaneous communications from UAVs to the BS are orthogonal in the frequency domain, then the communication link between UAV $i$ and the BS can be represented as
\begin{equation}
    \label{eq:channel}
    h_{i}=\sqrt{\beta_0} d_{i}^{-\frac{\alpha_{i}}{2}},\; \forall i=1,\ldots,N_U,
\end{equation}
where $d_{i}$ denotes the 3D distance between UAV $i$ and the BS, $\beta_0$ is the reference channel gain, and $\alpha_{i}$ is the path-loss component expressed as
\begin{equation}
    \label{eq:pathloss}
    \alpha_{i}=\frac{a_1}{1+a_4 \exp\left({a_3 \left( \theta_{i} - a_4 \right) }\right)}+a_2, \; \forall i =1,\ldots,N_U,
\end{equation}
where $\theta_{i}$ is the elevation angle between the BS and UAV $i$, and $\{a_1, \ldots,a_n\}$ are parametric constants related to the urban environment \cite{Hourani} \footnote{Different air-to-ground channel models can be considered. However, their impact is expected to be similar to that of the considered channel model since variations in channel quality would lead to the same effect from the FL perspective, e.g., slower FL convergence, lower FL accuracy, and higher UAV power consumption.}. Subsequently, the achievable uplink and downlink data rates for FL data exchange can be, respectively, given by
\begin{equation}
    R_{i,BS}= B_i \log_2 \left(1 + \frac{P_i |h_{i}|^2}{B_i \sigma^2} \right),\; \forall i =1,\ldots,N_U,
\end{equation}
and 
\begin{equation}
    R_{BS,i}= B \log_2 \left(1 + \frac{P_{BS} |h_{i}|^2}{B \sigma^2} \right),\; \forall i =1,\ldots,N_U,
\end{equation}
where $B_i$ (resp. $B$) is the bandwidth allocated to the transmission of UAV $i$ (resp. the BS), $P_i$ (resp. $P_{BS}$) is the transmit power of UAV $i$ (resp. the BS), and $\sigma^2$ is the unitary power of the additive white Gaussian noise (AWGN).
}
Hence, the required times for FL updates' upload over the uplink, and download over the downlink, can be respectively expressed by 
\begin{equation}
\label{eq:tx1}
    t_{i,BS}={V z}/{R_{i,BS}},
\end{equation}
and 
\begin{equation}
\label{eq:tx2}
    t_{BS,i}={V z}/{R_{BS,i}},
\end{equation}
where $V$ is the number of NN model parameters and $z$ is the size of a single model parameter, e.g., a NN weight. 
Consequently, the overall duration of global FL round $r$ at the EC server is expressed as follows:

\begin{equation}
    t_{r}=\max_{i =1,\ldots,P_r} \left(t_{i,r}+t_{i,BS}\right) + t_{ag},
\end{equation}
where $t_{ag}$ is the time taken by the EC server to perform aggregation on the participants' weight updates. Moreover, since $t_{BS,i}$ is negligible, thus it is eliminated from $t_r$.


\subsection{EC Server and UAVs' Security Profiles}
In our system, we assume that the EC server is benign, i.e., is not prone to cyberattacks and do respond correctly to FL operations. Regarding the UAV participants, we consider two security profiles,
namely reliable UAVs and unreliable UAVs. Expressly, we assume that: 
\begin{itemize}
    \item A reliable UAV $i$ is an honest participant that accurately reports its model parameter values, i.e., it uploads non-tampered vector weights $w_{i,r}$ to the EC server in round $r$.
    \item An unreliable UAV would demonstrate unreliable behavior in different manners. The first manner consists of dropping out, intentionally or not, from FL training after it started. The second manner is to submit poisonous model parameters to the EC server. This situation happens when an attacker takes control of one or several UAVs, makes them malicious, and obliges them to alter their own model parameters, either trained or transmitted ones. To align with practical systems, we assume that an attacker cannot corrupt more than 50\% of the available UAV participants. 
\end{itemize}
    
    

\section{Problem formulation and proposed solution}
\label{sec:problem-formulation}
\subsection{Problem Formulation}
In order to design a resilient FL system, we need to address the negative impact of statistical and system heterogeneity on the accuracy metric, denoted $A_{i,r}$, of each UAV participant $i$ in round $r$. Then, we address the security and reliability of the FL process in the presence of  unreliable UAV participants.  Formally, let $\textbf{x}_r=[x_{1,r},\ldots,x_{N_U,r}]$ be the vector of binary variables that indicate the participation or not of UAVs in FL round $r$, while $\zeta$ is a deadline for local training and update weights transmission.
We assume that among the participating $N_U$ UAVs, there exist four sets, namely $\mathcal{P}_r^h$, $\mathcal{P}_r^s$, $\mathcal{P}_r^d$, and $\mathcal{P}_r^m$, of sizes $P_r^h$, $P_r^s$, $P_r^d$, and $P_r^m$, respectively, such that $N_U=P_r^h+P_r^s+P_r^d+P_r^m$. These sets correspond respectively to honest (i.e., non-straggler, non-dropout, and non-malicious), stragglers, dropouts, and malicious participants\footnote{For the sake of simplicity we assume that $\mathcal{P}_r^h$, $\mathcal{P}_r^s$, $\mathcal{P}_r^d$, and $\mathcal{P}_r^m$ are distinct sets.}. 
Our objective consists of maximizing the accuracy by aggregating the update weights from $\mathcal{P}_r^h$ only. Let $\mathcal{\tilde{P}}_r^h$ be the decided set of participating UAVs in the aggregation process, based on the $\mathbf{x}_r$ strategy, i.e., $|\tilde{\mathcal{P}}_r^h|=\sum_{i=1}^{N_U}x_{i,r}$.
Subsequently, the related problem can be formulated as follows:



\begin{subequations}
	\begin{align}
	\label{p1}
	\max\limits_{\mathbf{x}_r, \zeta} & \quad \sum_{i=1}^{N_{U}} x_{i,r}.A_{i,r}   
	 \tag{P1} \\
	\label{cc1}
	\text{s.t.}\quad 
        & t_{i,r}+t_{i,BS} \leq \zeta, \forall i \in \mathcal{\tilde{P}}_r^h,  \tag{P1.a} \\  
     &  0 \leq \sum_{i=1}^{N_U} x_{i,r} \leq N_U, \tag{P1.b} \\
    &  x_{i,r} \in \{0,1\},  \forall i=1,\ldots, N_{U}, \tag{P1.c}
	\end{align}
\end{subequations} 
where constraint (P1.a) indicates that local training and update weights transmission should not exceed a deadline $\zeta$. If $\zeta$ is exceeded, the concerned UAV participant is considered as a dropout.
Constraint (P1.b) ensures that the maximal number of participants does not exceed the number of available UAVs, while (P1.c) reflects the binary nature of the participant selection variable.


 
Problem (P1) can be proven NP-hard. 
Indeed, for a given $\zeta$, it aims to maximize a sum objective, while respecting a resource constraint (maximal number of UAV participants), (P1) can be reduced to a knapsack problem, which is already proven to be NP-hard \cite{knapsack}. 
Since it is intractable to directly find
the global optimum 
of (P1), we propose to solve it using a cascaded solution, where the first sub-solution eliminates the stragglers and sets $\zeta$, the second focuses on selecting non-dropouts from the remaining non-stragglers UAVs, while the third sub-solution filters out malicious participants from the resulting set of the second sub-solution. 
These sub-solutions are detailed in the subsections below.



\subsection{Proposed Non-stragglers Participant Selection}
\begin{figure}[!h]
\includegraphics[width=0.99\linewidth]{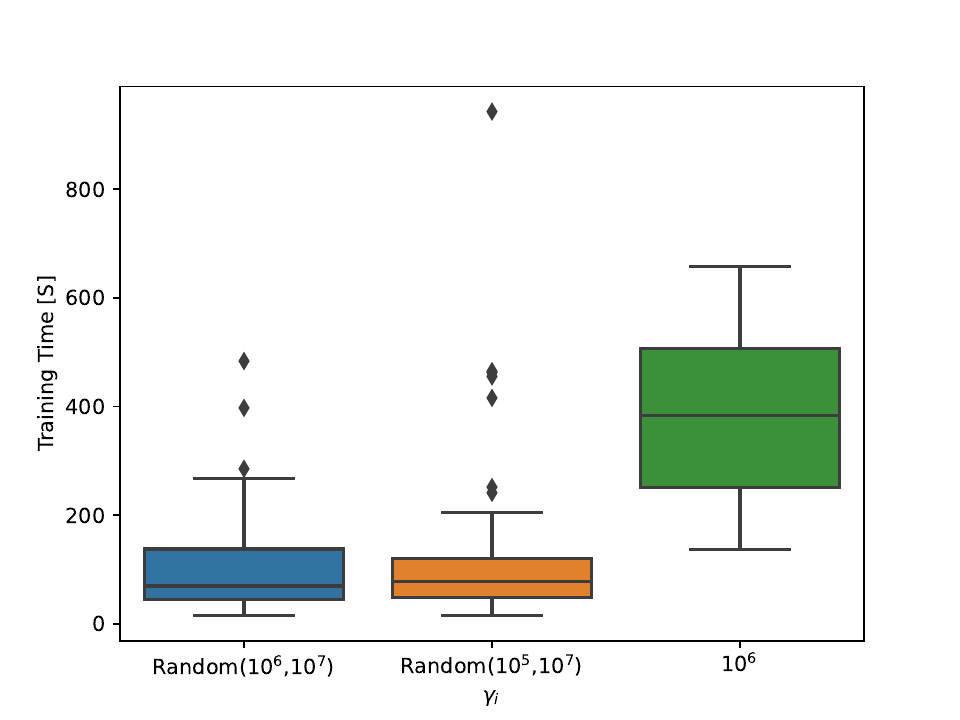}
\caption{Distribution of training round time for different UAVs' CPU frequencies.}
\label{outliers}
\end{figure}

Most research in the literature has focused on achieving a higher convergence rate for client scheduling by minimizing the number of communication rounds required to reach a specific level of global accuracy.
However, this approach does not minimize the elapsed round time during training.
Indeed, training time, $t_{i,r}$, is constrained by stragglers who would extend the training round time, due for instance to limited CPU frequency speed $\gamma_i$\footnote{To be noted that the EC server can estimate the UAVs' CPU information through the processing of a standardized training task on them and tracking the time required to complete it.}. In addition, 
participants may be training different numbers or sizes of data samples, due to the statistical heterogeneity, thus affecting the training round time.
In Fig. \ref{outliers}, we illustrate an example of the system heterogeneity's effect, in terms of $t_{i,r}$, and as a function of UAVs' CPU frequency ranges of $\gamma_i$.
As it can be seen, having UAVs with a large range of CPU capabilities, e.g., $\gamma_i \in [10^5, 10^7]$ Hz, resulted in the highest number of outliers or stragglers (black diamonds). 

Stragglers can be eliminated from the participants' set $\mathcal{P}_r$ by applying the
interquartile range (IQR) \cite{iqr}, which is a popular statistical method
that detects outliers. Our choice of IQR is motivated by its high robustness in skewed data. IQR value is calculated as the difference between quartiles $Q3$ (i.e., third quartile of the data or upper 75\% of data), and $Q1$ (i.e., lower 25\% of data).
Differently from the classic IQR method, we define our decision range as the upper bound of the accepted data points, and we do not consider the lower bound. Hence, any data point that has $t_{i,r}$ larger than the upper bound, defined as 
\begin{equation}
\label{eq:iqr}
IQR_{\rm up}=Q3 + \nu \times IQR, 
\end{equation}
where $\nu$ is a scaling parameter, where $\nu >0$. Thus, the participants' filtering engine eliminates stragglers from $\mathcal{P}_r$ and generates a novel set of UAVs, denoted $\mathcal{P}_r^{ns}$. 
Subsequently, we define by 
\begin{equation}
\label{eq:zeta}
\zeta=\frac{2}{|\mathcal{P}_r^{ns}|} \sum_{i \in \mathcal{P}_r^{ns}} t_{i,r},    
\end{equation}
the deadline for local training and update weights transmission. For the sake of simplicity, the value of $\zeta$ is chosen conservatively (twice the average local training time of non-stragglers), since transmission time is typically one or several orders of magnitudes lower than training time. 

Within eq. (\ref{eq:iqr}), the scaling factor $\nu$ is a constant multiplier that is applied to the IQR in order to determine the outliers' threshold. Actually, the value of $\nu$ to be used significantly depends on the application type and datasets properties. Nevertheless, most applications in the literature opted for a scaling factor of 1.5, as it is agreed to be a typical acceptable value \cite{10.2307/2683468,7030658}. In any case, if the training time showcases extreme values or high variability, a higher scaling factor would be preferable. In contrast, a small scaling factor would be suitable to prevent false positives if the training times are homogeneous.
In addition, using a large $\nu$ would lead to an aggressive straggler elimination strategy, which might affect the training accuracy, whereas a small $\nu$ value leads to a more conservative straggler elimination, and probably to longer training times.

\subsection{Proposed Non-dropouts Participant Selection}
\label{s2}

Given the high mobility and fragility of UAV systems,
unsteady communications or equipment failure may occur, which will lead to losing the client during the training round, i.e., dropping out of the FL process. As a result, the convergence and performance of the FL under partial participation mechanism is significantly impacted by the quantity of aggregated models. According to \cite{McMahan2017CommunicationEfficientLO}, increasing the number of aggregated local models improves the accuracy performance. However, in a realistic system, not all selected participants would succeed in training and uploading their updates within the fixed deadline. Consequently, the reliability of participants is vital to ensure convergence to an enhanced accuracy.


To mitigate dropouts and maintain resilient operations, we define here a reliability score, denoted by $\Lambda_{i,r}$ for UAV $i \in \mathcal{P}_r^{ns}$, which provides an assessment of the participant's behavior within the FL process. Specifically, a UAV uploading its model parameters within $\zeta$ is rewarded, while those failing, i.e., dropping out of the training round, are penalized. The reliability score is expressed by 
\begin{equation}
\label{eq:reliability}
    \Lambda_{i,r}= \Lambda_{i,r-1}+ {g}, \; \forall i \in \mathcal{P}_r^{ns},
\end{equation}
where 
$\Lambda_{i,r-1}$ is the obtained reliability score for the previous round $r-1$, and
\begin{equation}
    g = \begin{cases}
0, &\text{if UAV $i$ is asked to participate in round $r$} \\ & \text{but did not respond, or started responding} \\ 
& \text{but got disconnected before completing,}\\
1, & \text{if UAV $i$ participated and (P1.a) is satisfied}
\\
-1, &\text{otherwise.}
\end{cases}
\end{equation}
Assuming that a ``reliable'' (non-dropout) participant is defined by a value $\Lambda_{i,r}$ equal or greater than a reliability threshold $\rho$, then the set of potential participants can be reduced to $\mathcal{P}_r^{nd}= \{ i \in \mathcal{P}_r^{ns} | \Lambda_{i,r}\geq \rho \}$.  
Furthermore, in
order to avoid bias in the global model and fast battery exhaustion
of participants, we establish a client over-selection avoidance mechanism, where if $\Lambda_{i,r}$ reaches a pre-fixed maximum value $\rho_0$, it is reset to 0. 

In the particular case of a portion or all scores being lower than $\rho$, 
Based on the calculated $\Lambda_{i,r}$ values, the participant selection engine prioritizes the inclusion of $P_r-|\mathcal{P}_r^{nd}|$ high-score UAVs to $\mathcal{P}_{r}^{nd}$, which have $\Lambda_{i,r}<\rho$.

 \subsection{Proposed Non-malicious Aggregation}
 \label{s5}
  Following the previous steps to obtain a non-straggler non-dropout participants set $\mathcal{P}_r^{nd}$, we focus here on achieving non-malicious aggregation of update weights at the EC server.

As an input for our problem, we use the update weights inputs, denoted $w_r^{nd}$, of the participants' set $\mathcal{P}_r^{nd}$. The EC server evaluates  $w_{i,r}^{nd}$, $\forall i \in \mathcal{P}_r^{nd}$ to classify them as malicious (i.e., targeted or untargeted attack) or non-malicious. Then, malicious update weights are eliminated from FL aggregation. 

To do so, we start by characterizing the malicious attacks as follows. First, targeted attacks are easy to perform, e.g., erroneously classifying samples in a dataset, but difficult to detect since they subtly effect the global model. if the number of attackers is large, they typically demonstrate similar behaviours that are different from those with non-malicious intentions. Second, untargeted attacks demonstrate behaviours dissimilar from any other participant, thus their weight updates can be considered as outliers. A successful attack would eventually deviate the FL convergence direction towards a different one. By measuring the pairwise angular distance (PAD) between the participants' update weights, the desired convergence direction can be estimated. The use of PAD is preferred over other measures, such as the Euclidean distance, since malicious participants cannot alter the direction of a gradient without reducing their attack's efficacy. Hence, we propose here to calculate the PAD using the cosine similarity.     

Unlike \cite{https://doi.org/10.48550/arxiv.1712.01887} that calculated cosine similarity PAD on mini-batch gradients to estimate the direction of the true gradient, authors of \cite{
DBLP:journals/corr/abs-1910-01991} demonstrated that an approximate estimation of the true gradient's direction can be obtained by applying cosine similarity PAD on the FL's weight updates, under sufficiently smooth FL loss function and low learning rate conditions.
Consequently, we opt here for a similar approach as \cite{
DBLP:journals/corr/abs-1910-01991}, where the cosine similarities between weight updates, defined by $\kappa_{i,j}$, $\forall (i,j) \in \mathcal{P}_r^{nd} \times \mathcal{P}_r^{nd}$,  are calculated as 
\begin{equation}
\label{cosinesim}
    \kappa_{i,j}= \frac{< w_{i,r}, w_{j,r}>}{|| w_{i,r}|| ||w_{j,r}||},
\end{equation}
where $< \cdot, \cdot>$ is the Euclidean dot product and $||\cdot||$ is the magnitude's operator.

With the cosine similarity factors in hand, we transpose this information into a graph $\mathcal{G}$ having the elements of $\mathcal{P}_r^{nd}$ as vertices, and $\kappa_{i,j}$ as edges. According to the previous discussion, targeted attackers would behave similarly, while untargeted attackers can be seen as outliers. Backed with these assumptions, in addition to the ratio of attackers (f) being typically below half of participants in $\mathcal{P}_r^{nd}$, i.e., $\text{f}<50\%$, we can adopt a clustering approach to detect targeted and untargeted attackers in $\mathcal{G}$. For the sake of simplicity, we opt for the density-based spatial clustering of applications with noise (DBSCAN) to execute this task. The expected result is several clusters, where the largest one would correspond to the non-malicious participants, while smaller ones correspond to targeted attackers, and outliers to untargeted attackers. Subsequently, 
the EC server aggregates $w_{i,r}^{nd}$ from the non-malicious (largest) cluster only, denoted as $\tilde{\mathcal{P}}_r^h$.    

As it can be seen in the above subsections IV-B to IV-D, our solution consists of three cascaded sub-solutions, where the output of one sub-solution is the input of the next one. We summarize its operation in Algorithm \ref{Alg1}.

\begin{algorithm}{
  \algsetup{linenosize=\small}
  \small
		\SetAlgoLined
		\textbf{Input:} number of UAVs $N_U$, minimum number of participants $P_{r}$, number of global rounds $r_0$ \;			
		{\textbf{Steps:}}\\
        EC server initializes $w_0$ \; 
        Randomly assign $\Lambda_{i,1} \in [0,9]$, $\forall \; UAV_i \in \{1,\ldots,N_U\}$ \; 
        
        \For{round $r=1 \ldots r_0$}{
        /* Filter 1: Remove stragglers */ 
        \newline
        Calculate $IQR_{up}$ using eq.(\ref{eq:iqr}) \; 
        \For{each UAV$_i$ in $\{1,\ldots, N_{U}\}$}{
        \If{$t_{i,r} < IQR_{up} $}{
		$ \mathcal{P}_r^{ns}\leftarrow  \mathcal{P}_r^{ns} \cup \{UAV_i\};$ 
         }
  }
        Calculate $\zeta$ using eq.(\ref{eq:zeta}) \;
        /* Filter 2: Remove dropouts */ 
        \newline
         \For{each UAV$_i$ in ${P}_r^{ns}$}{
        \If{$\Lambda_{i,r} \geq \rho_0$} {
       $ \Lambda_{i,r} \leftarrow  0;$ 
       } 
       \If{$\Lambda_{i,r} \geq \rho$} {
       $ \mathcal{P}_r^{nd}\leftarrow  \mathcal{P}_r^{nd} \cup \{UAV_i\}$ \; 
       } 
       
 }

\If{$|P_r^{nd}| < P_r$} {
       Sort $\mathcal{P}_r^{ns} \backslash \mathcal{P}_r^{nd}$ in descending order w.r.t. $\Lambda_{i,r}$ \; 
       Add the first $P_r-|\mathcal{P}_r^{nd}|$ UAVs from $\mathcal{P}_r^{ns} \backslash \mathcal{P}_r^{nd}$ to $\mathcal{P}_r^{nd}$ (with highest $\Lambda_{i,r}$) \; 
       }

 /* Filter 3:Remove malicious participants */ \\
  \For{each UAV$_i$ in ${P}_r^{nd}$}{
             
Update local model using eq.(\ref{eq:SGD}) for $T_e$ local iterations\; 
        \For{each UAV$_j$ in $\mathcal{P}_r^{nd}$ where $j \geq i+1$}{
         Calculate and store cosine similarity $\kappa_{i,j}$ using eq.(\ref{cosinesim}) \;    
         } 
         } 
Apply DBSCAN and obtain clusters sorted from the largest to the smallest as $\mathcal{C}_r=\{\mathcal{C}_1,\ldots,\mathcal{C}_{|\mathcal{C}_r|}\}$ \;
  Set  $\tilde{\mathcal{P}}_r^h= \mathcal{C}_1$ \; 
   Update $w_{r}$ on $\tilde{\mathcal{P}}_r^h$ using eq.(\ref{eq:globalmodel}) \;
   Update $\Lambda_{i,r+1}$ using eq.(\ref{eq:reliability}), $\forall \; UAV_i \in \{ 1,\ldots,N_U \}$ \;
  }
  \textbf{Output:} $w_{r_0}$ global model after $r_0$ rounds of training \;
\caption{UAV reliable participation}
\label{Alg1}
}
\end{algorithm}


\section{Experiments and Results}
\label{sec: experiment}
\subsection{Experiment Setup}

We consider an FL system with a single BS/EC server and $N_U=50$ UAVs within its coverage area. The UAVs' computing frequencies $\gamma_i$, $\forall i=1,\ldots,N_U$, are randomly sampled from the range $[10^6,10^8]$ Hz. To filter out stragglers, we set $\nu=1.5$ \cite{10.2307/2683468,7030658}. Also, $\rho$ is set to -5 and $\rho_0$ to 10. These values were determined numerically through experimentation.
We assume that $\kappa=7 \times \; 10^4$ CPU cycles, $P_i=0.28$ Watt, $\forall i=1,\ldots,N_U$. Additionally, we have set up a network configuration wherein the model weights are transmitted from the FL client to the server with a total delay ranging between 20 ms and 200 ms, using a total bandwidth of 10 MHz.
We test our solution's effectiveness on an image classification FL task, where the dataset is split into a training set and a testing set at ratios 80\% and 20\%, respectively. We run our defense approach on two popular datasets, namely MNIST and CIFAR-10, where MNIST consists of black and white handwritten digits of 60,000 training samples with size $28 \times 28$ pixels \cite{deng2012mnist}, while CIFAR-10 is composed of colored pictures with size $32 \times 32$ pixels, including 50,000 training images and 10,000 testing images \cite{cifar10}. Both datasets are divided into 10 classes, for numbers from ``0'' to ``9'' in MNIST, and for object and animal classes in CIFAR-10. We assume that each UAV acan have a maximum number of $|M_r|=1300$ samples.
    To take into account dataset non-IIDness in our experiment, we use different distribution methods. To build the first data distribution, called ``Distribution 1'', we use random sampling with the notion of ``$k\%$ non-IID'', where $k$ measures the degree of non-IIDness in percentage. Accordingly, by setting $k=80$, 80\% of the dataset is dispersed randomly between the UAV FL clients while the remaining 20\% is equally distributed among them. This data distribution strategy leads to unbalanced local datasets. For the second distribution, called ``Distribution 2'', we intentionally spread the data such that each UAV has data issued from $N_c=2$ classes only. To be noted that both distributions 1 and 2 reflect two different strategies of data non-IIDness. This distributions design task is executed only once, and the resulting local datasets will be used in the remaining of this section.

In our MNIST experiment, we adopt at each FL participant the LeNet-5 convolutional neural network (CNN) architecture \cite{726791}, while we adopt the VGG11 architecture for CIFAR-10-related system \cite{vgg11}. The latter is composed of eight CNN layers and three fully connected layers.
We set the mini-batch size to 64, the number of local iterations for each round $T_e=10$, and we use the vanilla stochastic gradient descent optimizer. Moreover, we select the minimum number of participants $P_r=5$, 
$r_0=200$ as the number of FL rounds, $0.01$ as the learning rate, and the cross-entropy loss as the loss function. 
Finally, the hyperparameter $\epsilon$ for DBSCAN clustering is set to $\epsilon=0.02$ in the MNIST-related system, and to $\epsilon=0.06$ for the CIFAR-10 related system.


Here, we evaluate our solution's performance in two different settings. In the first setting, we assume statistical and system heterogeneous environments using different data distributions, where stragglers and dropouts can exist, but no attackers, i.e., the ratio of attackers is $f=0\%$. 
In the second setting, we consider $f>0\%$, with different targeted and untargeted attack scenarios. 
The experimented targeted and untargeted attacks are chosen as the label flipping attack and additive noise attack, respectively, which are detailed as follows:
\begin{itemize}
    \item \textbf{Label flipping attack:} It attempts to modify the behavior of an ML model on a subset of data samples while retaining the primary model performance over the whole testing dataset. Specifically, attackers attempt to compel the model to identify samples with the label ``5'' as images with label ``3'', as shown in Fig. \ref{fig:syst} (Top-left rectangle). 
    \item \textbf{Additive noise attack:} During this attack, a malicious participant contributes with Gaussian noise in its local model updates in order to degrade the global model's accuracy and prevent its convergence, as illustrated in Fig. \ref{fig:syst} (Top-middle rectangle).
\end{itemize}

The performance metrics evaluated in our system are the FL accuracy in the absence of malicious participants ($f=0\%$), while under attacks ($f >0\%$), we evaluate the attack success rate, denoted as ASR, and the false positive (FP) and false negative (FN) ratios of detected attackers. In a targeted attack scenario, ASR$_{ta}$ is defined as 
the ratio of the number of altered dataset samples (i.e., those with their label flipped from `5' to `3') incorrectly classified (i.e., classified to anything but `5') to the total number of samples within the dataset.
It can be written as
\begin{equation}
\text{ASR}_{ta}=|\mathcal{S}_{ta}|/|\mathcal{S}|,
\end{equation}
where $|\mathcal{S}_{ta}|$ (resp. $|S|$) is the size of incorrectly classified data from the attacked class dataset (resp. size of dataset). 
Similarly, ASR$_{ua}$ for an untargeted attack can be expressed by
\begin{equation}
    \text{ASR}_{ua}=|A_{1}-A_2|/A_1,
\end{equation}
where $A_1$ and $A_2$ are the measured accuracy values under the scenarios of no attacks and an untargeted attack, respectively.  
Moreover, FP denotes the ratio of honest participants wrongfully identified as attackers, while FN refers to the ratio of non-detected attackers.




\begin{figure}[!h]
  \centering

  \begin{minipage}{.85\linewidth}
    \centering
    \subcaptionbox{Distribution 1}
      {\includegraphics[trim={0.4cm 6.25cm 1.5cm 7.2cm},clip,width=\linewidth]{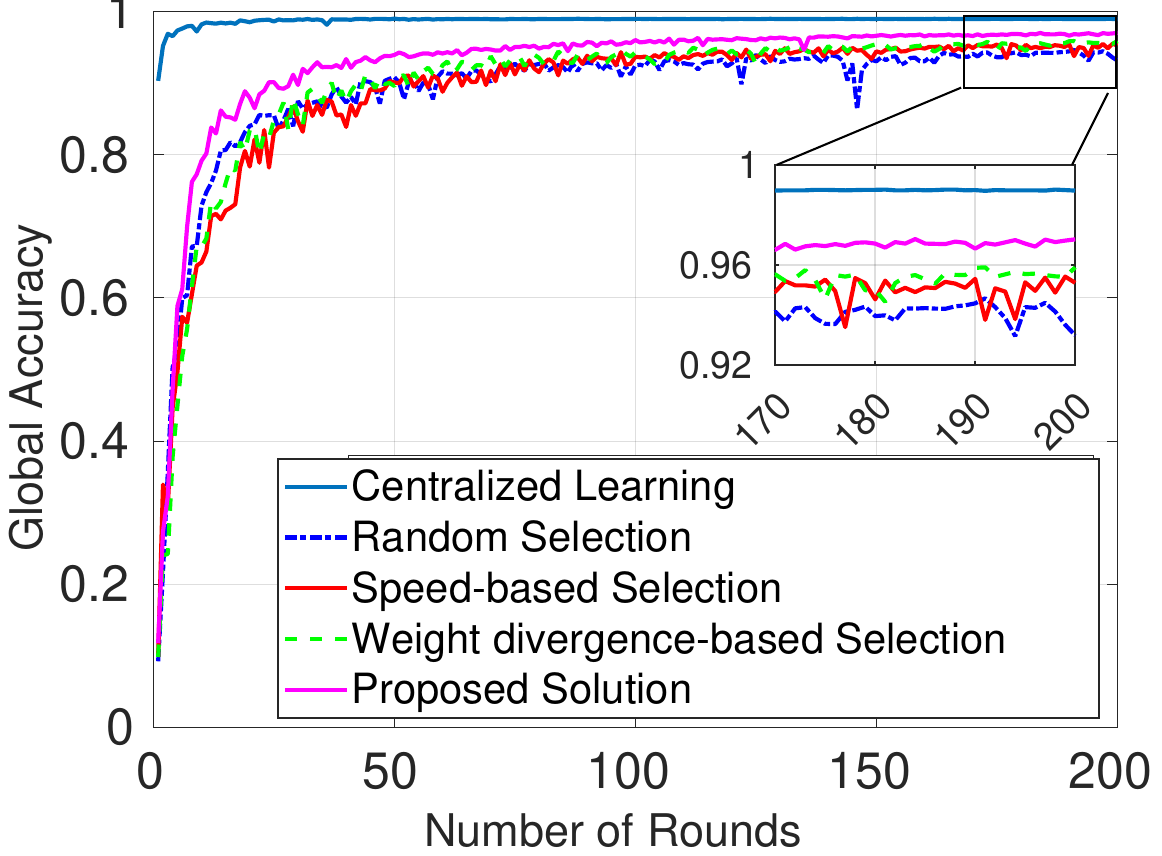}}
      

    \subcaptionbox{Distribution 2}
      {\includegraphics[trim={1.5cm 6.5cm 2.25cm 7.2cm},clip,width=\linewidth]{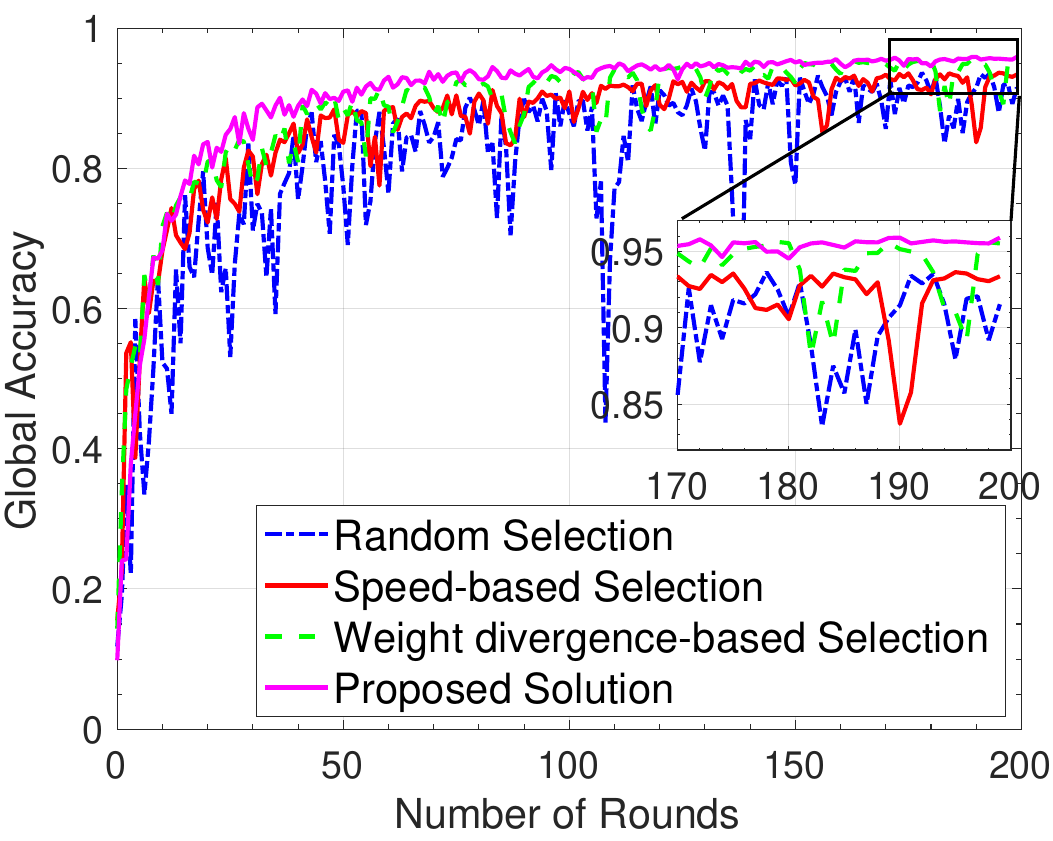}}

    \caption{Global accuracy vs. FL rounds under different data distributions (MNIST dataset).}
       \label{AccBenign}

  \end{minipage}
\end{figure}

\begin{figure}[!h]
  \centering

  \begin{minipage}{.85\linewidth}
    \centering
    \subcaptionbox{Distribution 1}
      {\includegraphics[trim={0.4cm 6cm 1.5cm 7.2cm},clip,width=\linewidth]{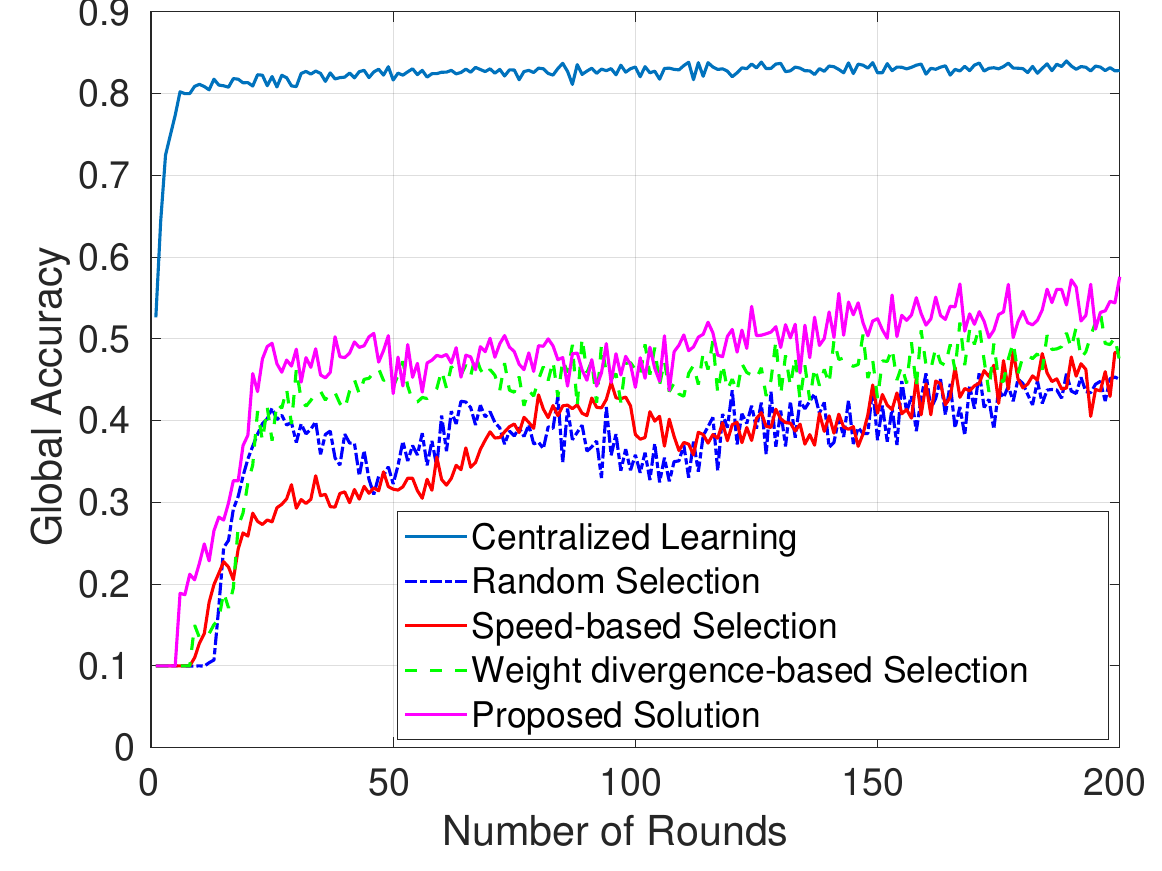}}
      

    \subcaptionbox{Distribution 2}
      {\includegraphics[trim={0cm 6cm 1cm 6cm},clip,width=\linewidth]{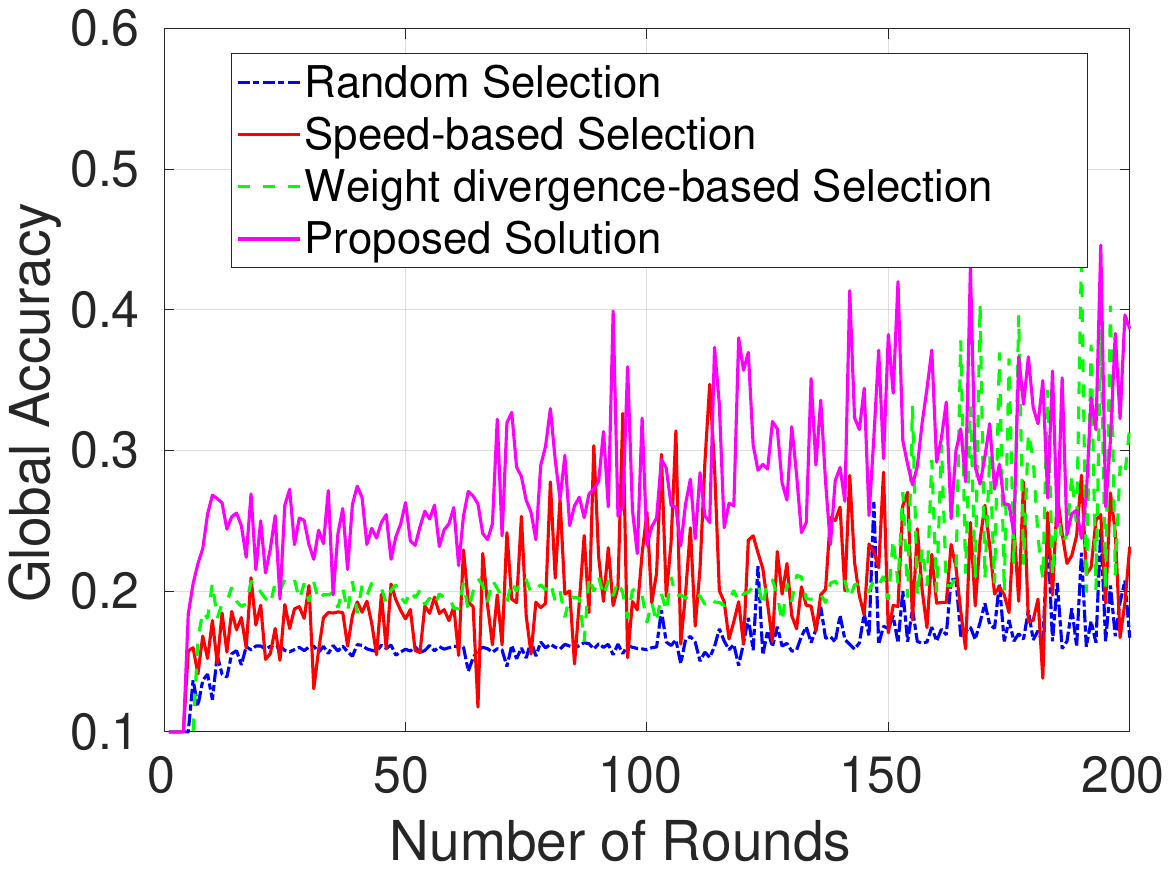}}

    \caption{Global accuracy vs. FL rounds under different data distributions (CIFAR-10 dataset).}
       \label{AccBenign2}

  \end{minipage}
\end{figure}

\begin{table}[!h]
\caption{Comparison of participants selection methods in terms of accuracy, dropout ratio, and average round time (MNIST dataset).}
    \centering
    \begin{tabular}{|c*{4}{c}|}

\hline Distribution & Selection Approach & $A_g$ & $\tau_{a}$ & $\chi_d$\\
\hline
 &Weight divergence-based &  95.90 \% &37.13& 0.15 \\
       \multirow{1}{*}{Distribution 1} & \makecell{Speed-based} &  95.28 \% &1.82& 0.11\\
          &Random & 93.18\% &41.85& 0.15\\
          
          &\textbf{Proposed solution} &  {97.02\%} & 7.31& 0.05\\
          \hline

           &Weight divergence-based & 94.12 \% &36.82& 0.17 \\
       
       \multirow{1}{*}{Distribution 2} &Speed-based &  93.37 \% &2.10& 0.10\\
          
          &Random & 91.54\% &38.35& 0.18\\
          
          &\textbf{Proposed solution} &  95.52\% & 5.52& 0.037\\
          \hline
          
    \end{tabular}
    
    \label{tab:1}
\end{table}
\begin{table}[!h]
\caption{Comparison of participants selection methods in terms of accuracy, dropout ratio, and average round time (CIFAR-10 dataset).}
    \centering
    \begin{tabular}{|c*{4}{c}|}

\hline Distribution & Selection Approach & $A_g$ & $\tau_{a}$ & $\chi_d$\\
\hline
 &Weight divergence-based & 48.16\% & 32.70 & 0.14 \\
       \multirow{1}{*}{Distribution 1} & \makecell{Speed-based} &  46.34\% & 2.32 & 0.1\\
          &Random & 45.03\% & 39.89 & 0.17\\
          
          &\textbf{Proposed solution} & 57.5\%  &  6.55 & 0.06\\
          \hline

           &Weight divergence-based & 31.71\% & 39.4 & 0.17 \\
       
       \multirow{1}{*}{Distribution 2} &Speed-based &  23.17\% &1.70& 0.12\\
          
          &Random & 16.70\% & 40.34& 0.16\\
          
          &\textbf{Proposed solution} & 38.80\%  & 7.65 & 0.05\\
          \hline
          
    \end{tabular}
    
    \label{tab:11}
\end{table}

\begin{figure}[!h]
  \centering
  \label{ASRUT}
  \begin{minipage}{.85\linewidth}
    \centering
    \subcaptionbox{$f$=10\%}
      {\includegraphics[trim={1.15cm 0.25cm 2.17cm 1cm},clip,width=\linewidth]{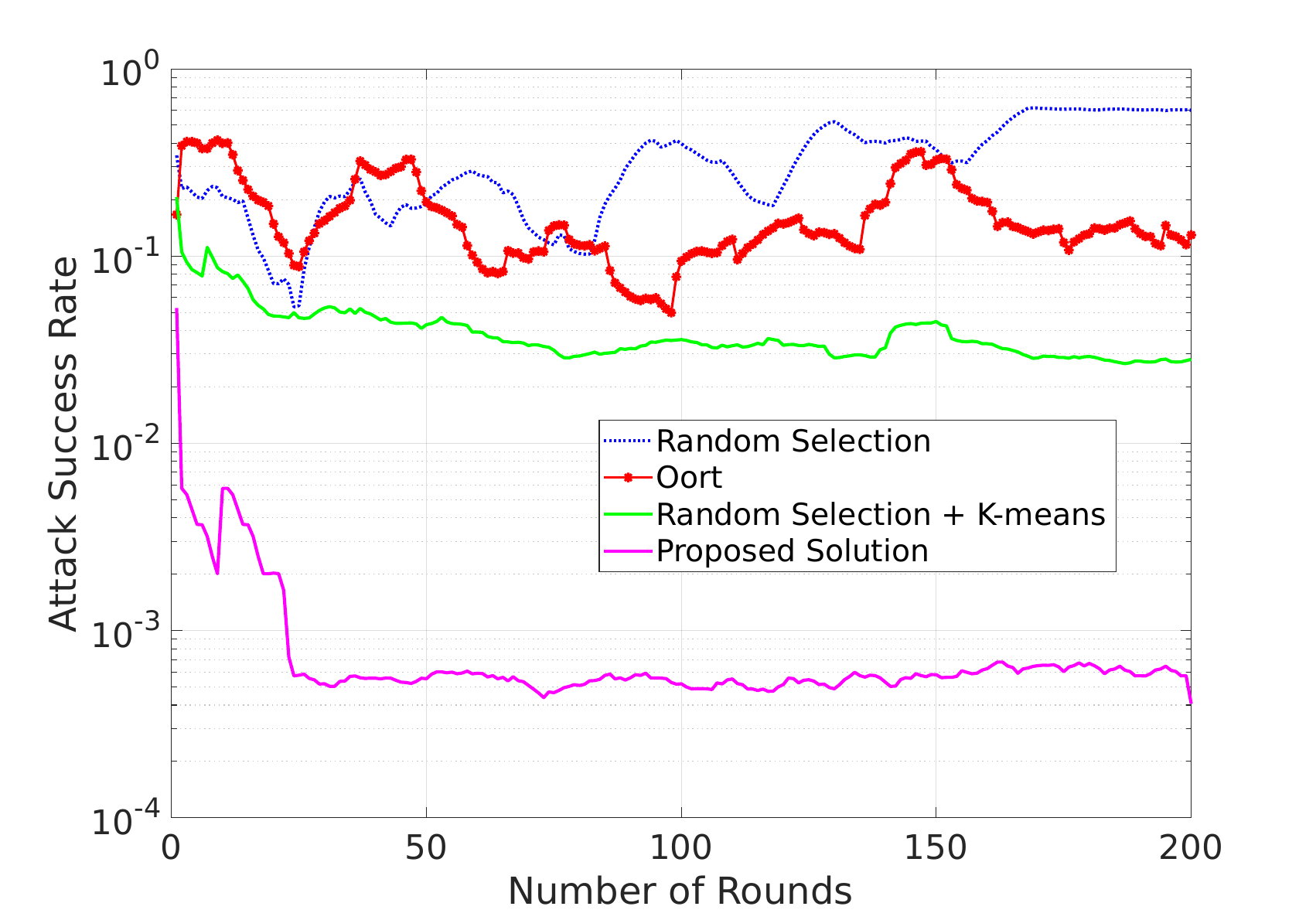}}
    \subcaptionbox{$f$=20\%}
      {\includegraphics[trim={1cm 6.5cm 2.25cm 7.2cm},clip,width=\linewidth]{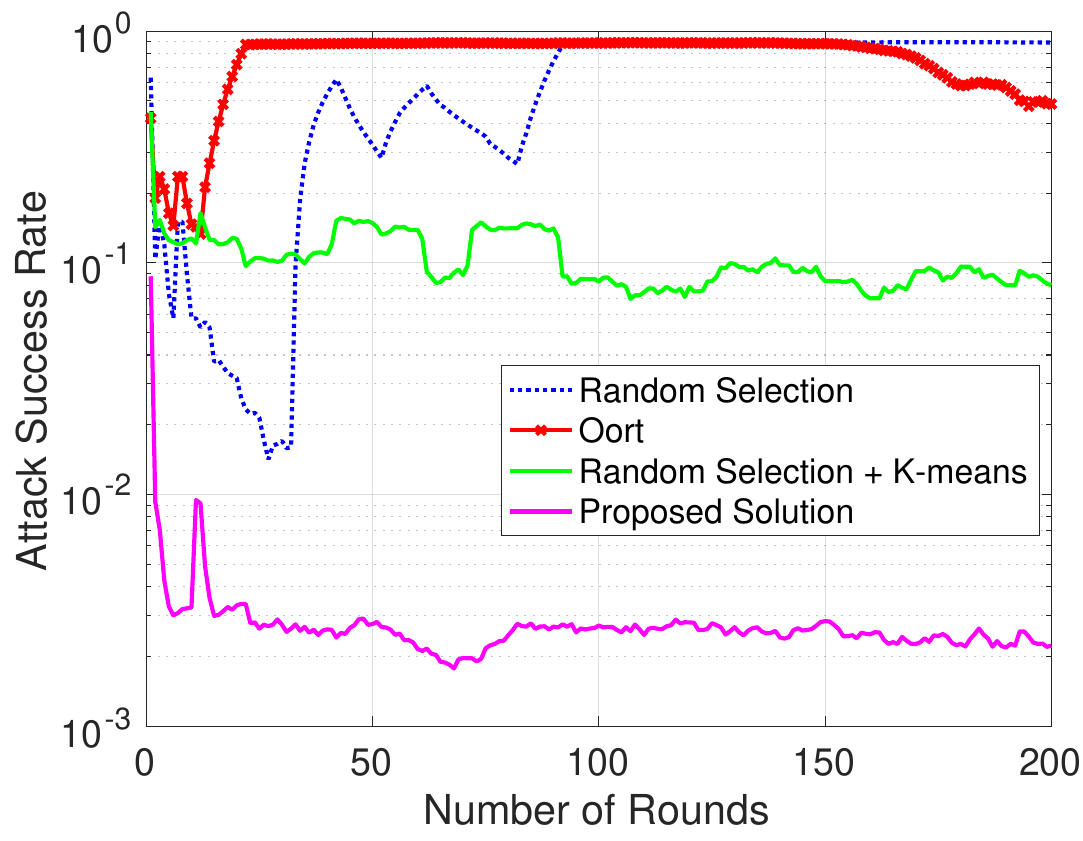}}
    \subcaptionbox{$f$=33\%}
      {\includegraphics[trim={1.15cm 0.25cm 2.17cm 1cm},clip,width=\linewidth]{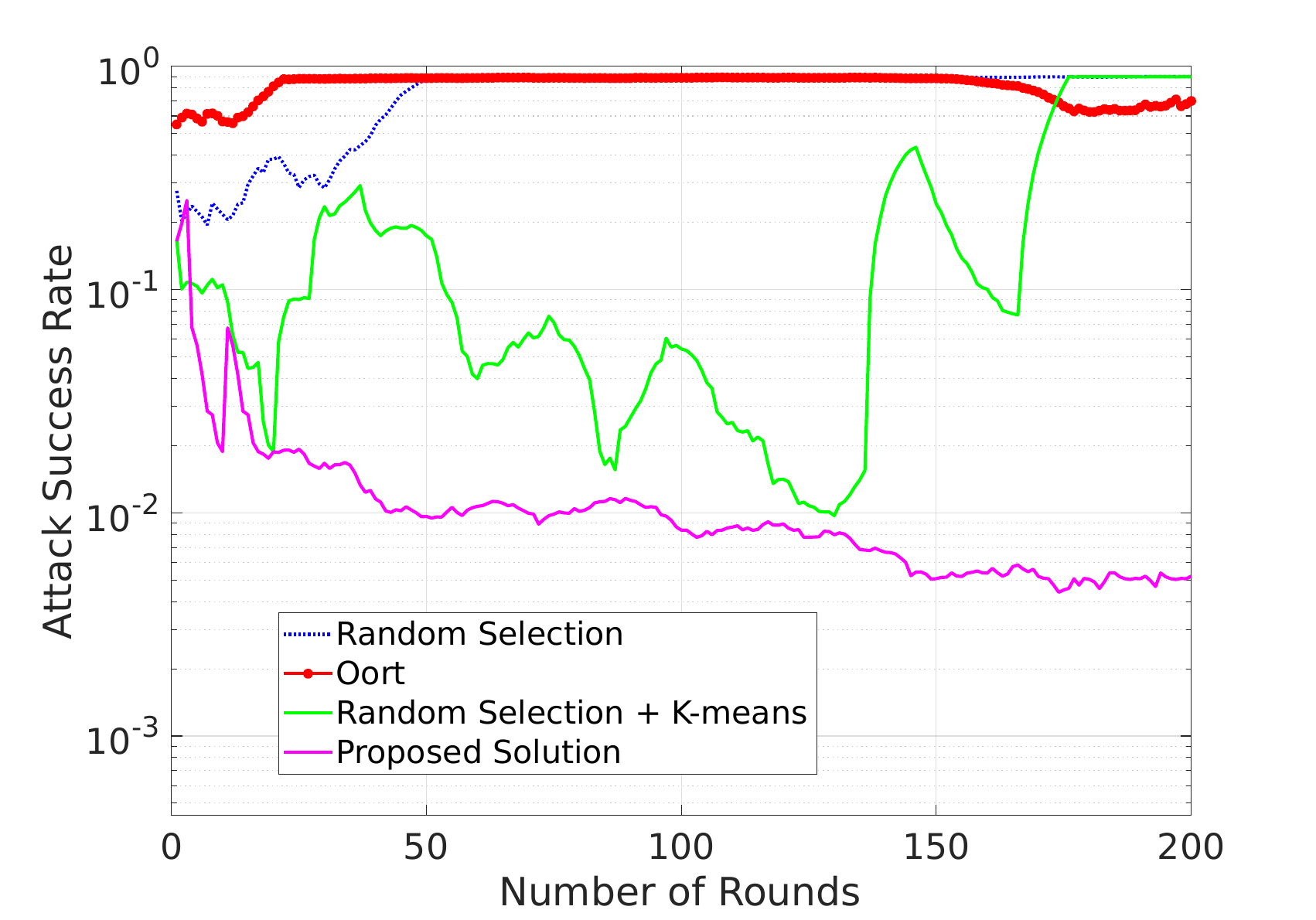}}
    \caption{Attack success rate vs. FL rounds under untargeted attack (MNIST dataset).}
  \end{minipage}
\end{figure}

\subsection{Results and Discussion}
\subsubsection{Setting 1 ($f=0$)}


We evaluate the global accuracy, denoted $A_g$, of our solution, and compare it to those of four baseline methods.  The first benchmark is the centralized learning approach, which is a model training technique characterized by the collection and processing of all data in a single server. 

The second baseline is the random selection scheme of participants, denoted ``Random Selection'', which operates conventionally with FL. The third baseline, named ``Speed-based Selection'', chooses the fastest participants in terms of data processing time. The forth approach called ``Weight divergence-based Selection'', defines the weight divergence metric as the distance between the weights of the local and global models, calculated as
\begin{equation}
\label{eq:weightdivergence}
    \hat{w}_{i,r}= \left|\frac{w_{i,r}- w_{r-1}}{w_{r-1}}\right|, \; \forall i =1,\ldots,N_U,
\end{equation}
and describes the update drift in non-IID datasets conditions \cite{9430906, 10.1007/978-3-030-10925-7_24}. Specifically, a large weight divergence value implies that the client's data has a high non-IIDness degree compared to other participants, thus it should take part in more upcoming rounds to reduce this gap and maximize its accuracy. Moreover, selecting participants with larger weight divergence values would prevent bias, which is a result of the global model ``drifting'' to a client's local optimizer \cite{article22}.

In Fig. \ref{AccBenign}a and Fig. \ref{AccBenign2}a, we present the global accuracy of the aforementioned participants selection methods, under Distribution 1 scenario, and using the MNIST and CIFAR-10 datasets, respectively.
Accordingly, centralized learning achieves the best accuracy performance of 98.9\% for the MNIST dataset and 82.4\% for the CIFAR-10 dataset after 50 epochs\footnote{Here, the number of rounds represents the number of epochs for centralized learning.}. In contrast, the remaining techniques achieve performances ranging between 90\% (random selection) and 96\% (proposed solution) for MNIST, while for CIFAR-10 their performances are between 45.03\% (random selection) and 57.5\% (proposed solution). The findings indicate that the convergence and performances of FL, even when optimized, are below those of centralized learning, as a result of data non-IIDness.\footnote{Given that centralized learning is computationally heavy and due to its very high performance gap with FL techniques, we decide to drop it in the remaining experiments.}
In any case, the proposed solution demonstrates its superiority over the FL baselines. 

The superior performance of our solution is mainly due to the careful participant selection that guarantees a small number of dropouts and a small round time, which is not the case with most baselines. Indeed, through the results of Table \ref{tab:1} and Table \ref{tab:11} (Distribution 1), we can see that our solution has the lowest dropout ratio $\chi_d$, and the second lowest average round time $\tau_a$. ``Speed-based Selection'' achieves the best $\tau_a$ performance since it always selects participants with low processing times, regardless of their dropout risk. 
The proposed solution offers a good trade-off between reliability and convergence speed while maximizing global accuracy.





\begin{figure}[!h]
  \centering
  \label{ASRUT-Cifar}
  \begin{minipage}{.85\linewidth}
    \centering  \subcaptionbox{$f$=10\%}
      {\includegraphics[trim={1cm 5cm 2.25cm 6.5cm},clip,width=\linewidth,height=170pt]{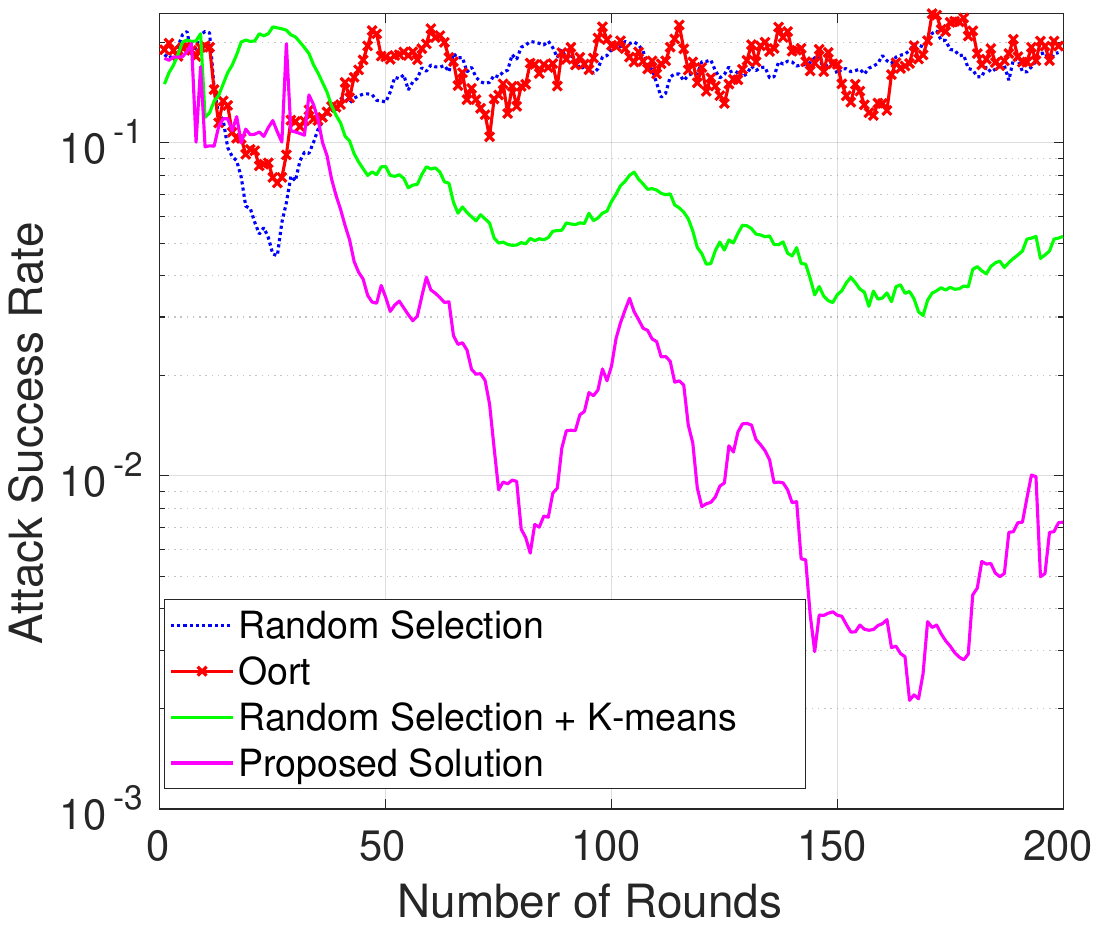}}
\subcaptionbox{$f$=20\%}
      {\includegraphics[trim={1cm 6.5cm 2.25cm 7.2cm},clip,width=\linewidth]{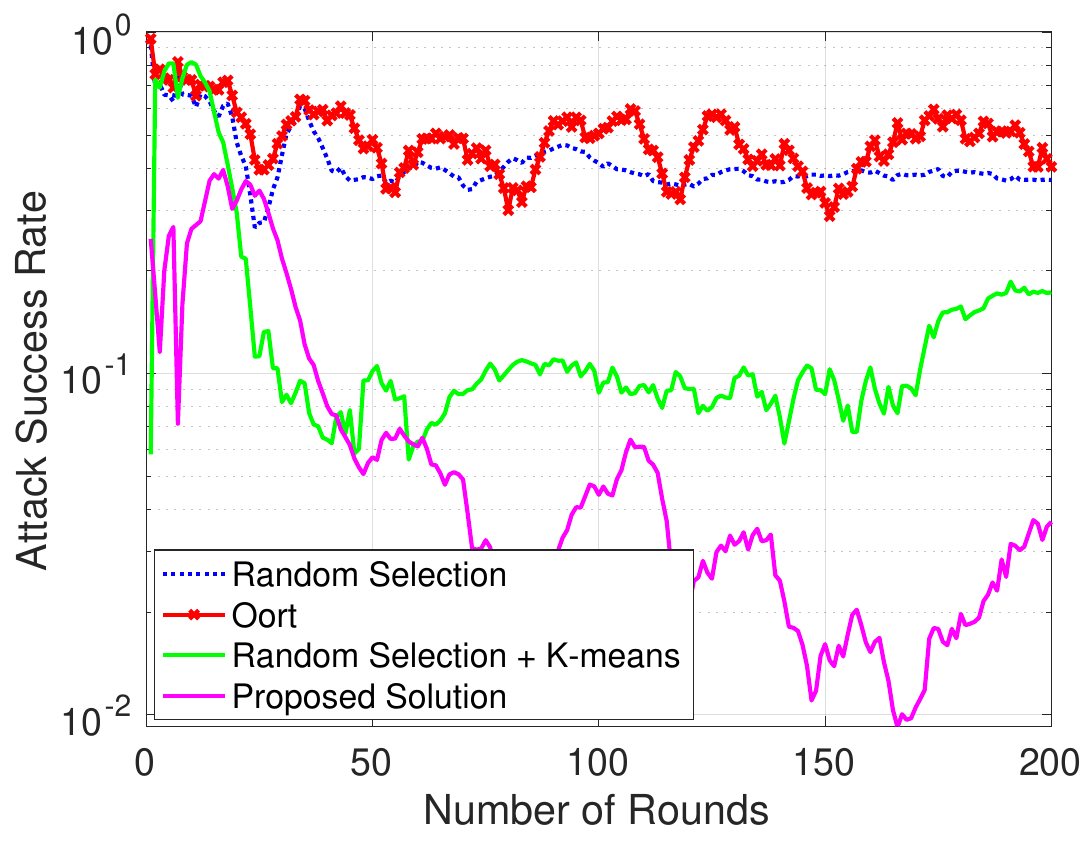}}
\subcaptionbox{$f$=33\%}
      {\includegraphics[trim={0cm 6.25cm 1cm 6cm},clip,width=\linewidth]{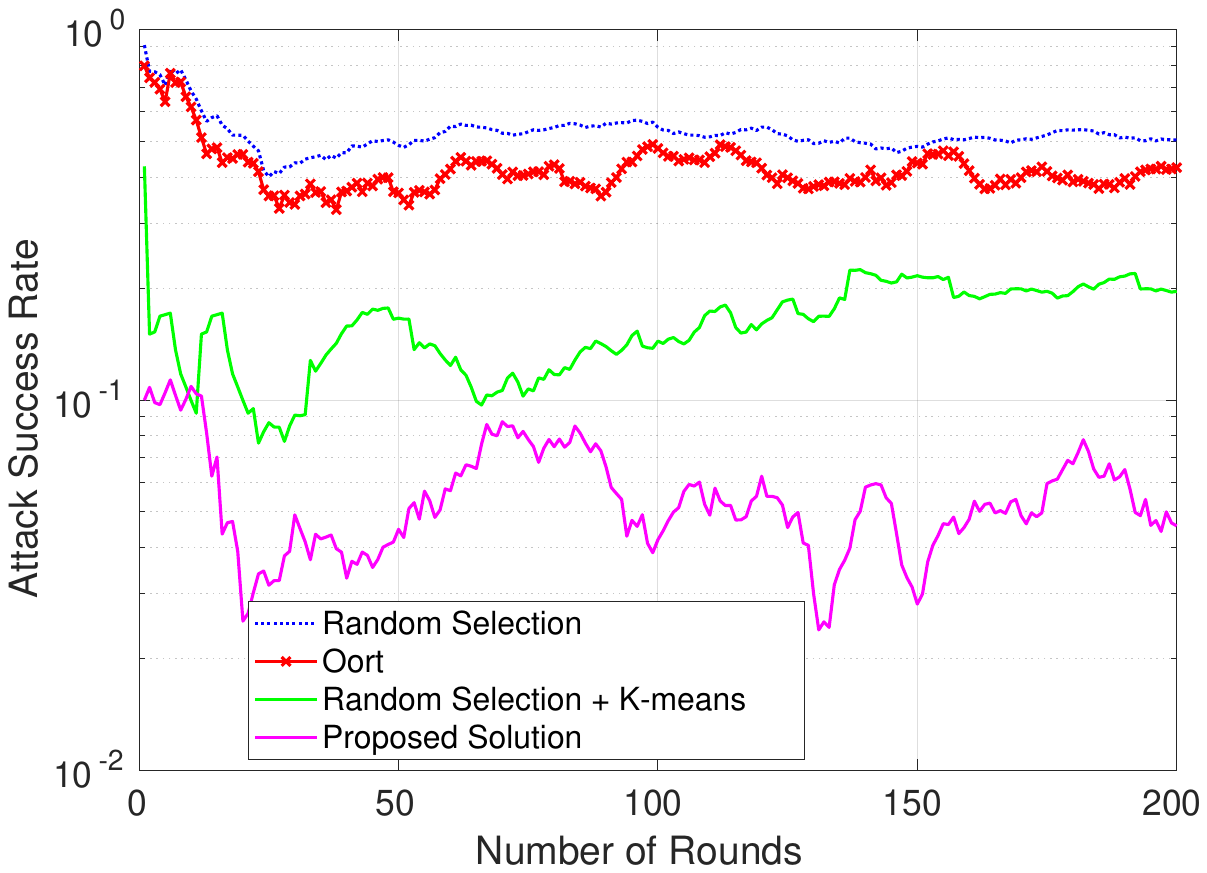}}
    \caption{Attack success rate vs. FL rounds under untargeted attack (CIFAR-10 dataset).}
  \end{minipage}
\end{figure}

In Fig. \ref{AccBenign}b Fig. \ref{AccBenign2}b, we extend the performance evaluation to Distribution 2. As noticed, the proposed solution is robust to highly unbalanced data distribution and achieves high accuracy, compared to baselines. Also, it convergences the fastest. 
Similarly to Distribution 1, the dropout ratio and average round time are the lowest and second lowest ones as shown in Table \ref{tab:1} and Table \ref{tab:11} (Distribution 2).


\subsubsection{Setting 2 ($f>0$)}
In this setting, and under the data assumption of Distribution 1, we compare our solution to three baselines. First, we use the previously defined ``Random Selection'' where malicious attackers filtering is in place. Then, we define ``Random Selection + K-means'', which adds to ``Random Selection'' a K-means \cite{1056489} based clustering defense mechanism to detect attackers when eq.(\ref{cosinesim}) is applied. The number of clusters is set to $C=2$, in order to distinguish between the non-malicious and malicious UAVs. Finally, the third baseline is ``Oort'' \cite{273723}, which relies on controlling participation to FL by limiting the number of participation rounds per UAV to 20.  
Experiments have been conducted  with attackers ratios (among available $N_U=50$ UAVs) $f \in \{10\%, 20\%, 33\% \}$.


    

\noindent
\textbf{\textit{Performance under untargeted attacks:}}
Under additive noise attacks (untargeted attacks), we evaluate in Fig. 6 and Fig. 7 the ASR$_{ua}$ performances of the proposed solution and baselines, given different $f$ ratios, and for the MNIST and CIFAR-10 systems, respectively. For any $f$, the proposed solution (pink lines) achieves the lowest, and thus the best, performance in terms of attack success rate. Looking into the subfigures in Fig. 6 (resp. Fig. 7), we notice that our solution succeeds in keeping the ASR$_{ua}$ below 1\% (resp. 10\%) for most of the time with MNIST (resp. CIFAR-10).
In contrast, all baselines' performances are worse than that of our method, for any $f$. Specifically, ``Random Selection'' (blue lines) and ``Oort'' (red lines) achieve the worst ASR$_{ua}$. Indeed, since ``Oort'' eliminates participants through rounds, at some point, it can improve the ASR performance. However, since this method does not clearly distinguish between non-malicious and malicious UAVs, over time, the UAV pool from which participants are selected is shrinking, thus leading to degraded performances.  
In the meanwhile, ``Random Selection + K-means'' demonstrates better results than the previous baselines, but still worse than our solution. Moreover, as $f$ increases, this method demonstrates less robustness.  
The poor performance of this method is due to the high FP and FN of attackers' detection using K-means, thus causing the non-detection of malicious UAVs and the elimination of non-malicious UAVs during training. In Tables \ref{tab:2} and \ref{tab:cifar}, we see that as $f$ increases, FP and FN deteriorate for the benchmarks, hence leading to worse results. Finally, the obtained ASR$_{ua}$ results for MNIST are better than those for CIFAR-10, for any method. Indeed, the CIFAR-10-related task is more complex, thus a stronger data non-IIDness is observed, leading to the cluster-based selective aggregation being less efficient.

\begin{figure}[!h]
  \centering

  \begin{minipage}{.85\linewidth}
  
    \centering
    \subcaptionbox{$f$=10\%}
      {\includegraphics[trim={1cm 6.5cm 2.25cm 7.2cm},clip,width=\linewidth]{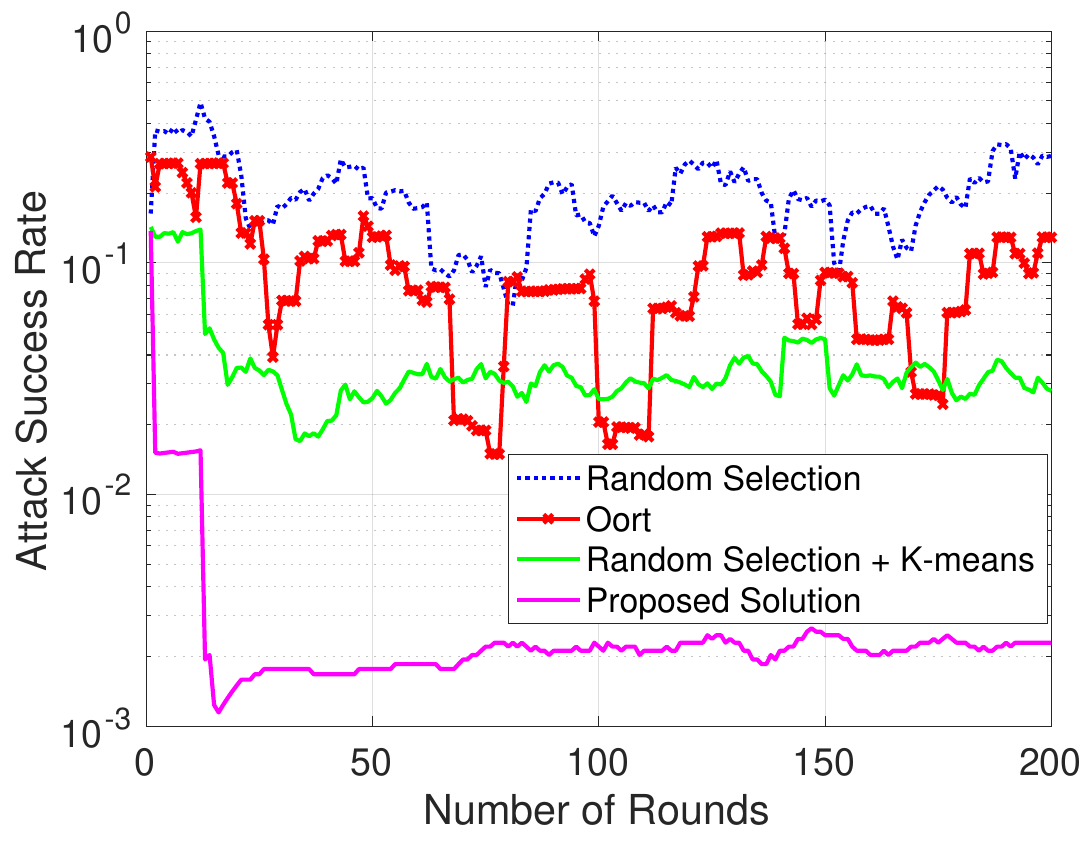}}

    \subcaptionbox{$f$=20\%}
      {\includegraphics[trim={0cm 6cm 1.5cm 6.5cm},clip,width=\linewidth]{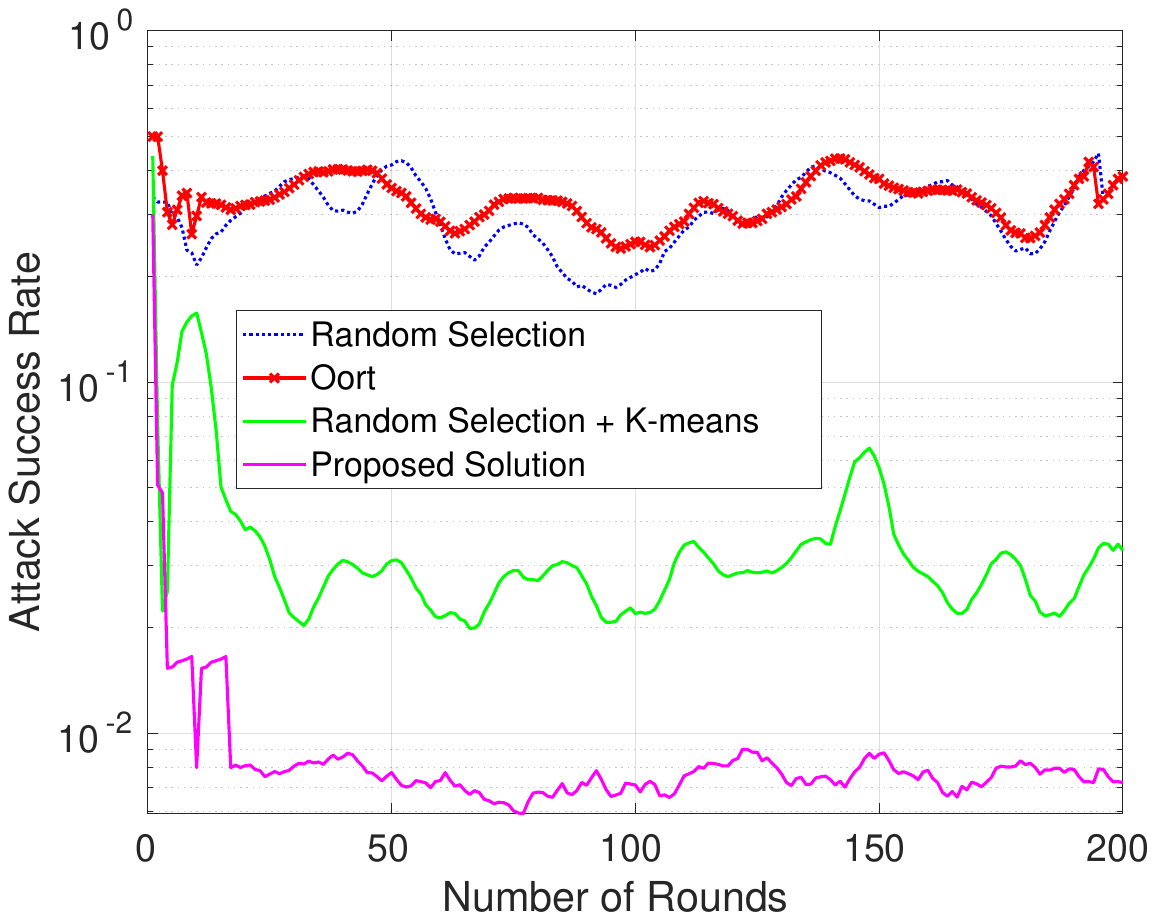}}

    \subcaptionbox{$f$=33\%}
      {\includegraphics[trim={0cm 6.4cm 1.2cm 6.5cm},clip,width=\linewidth]{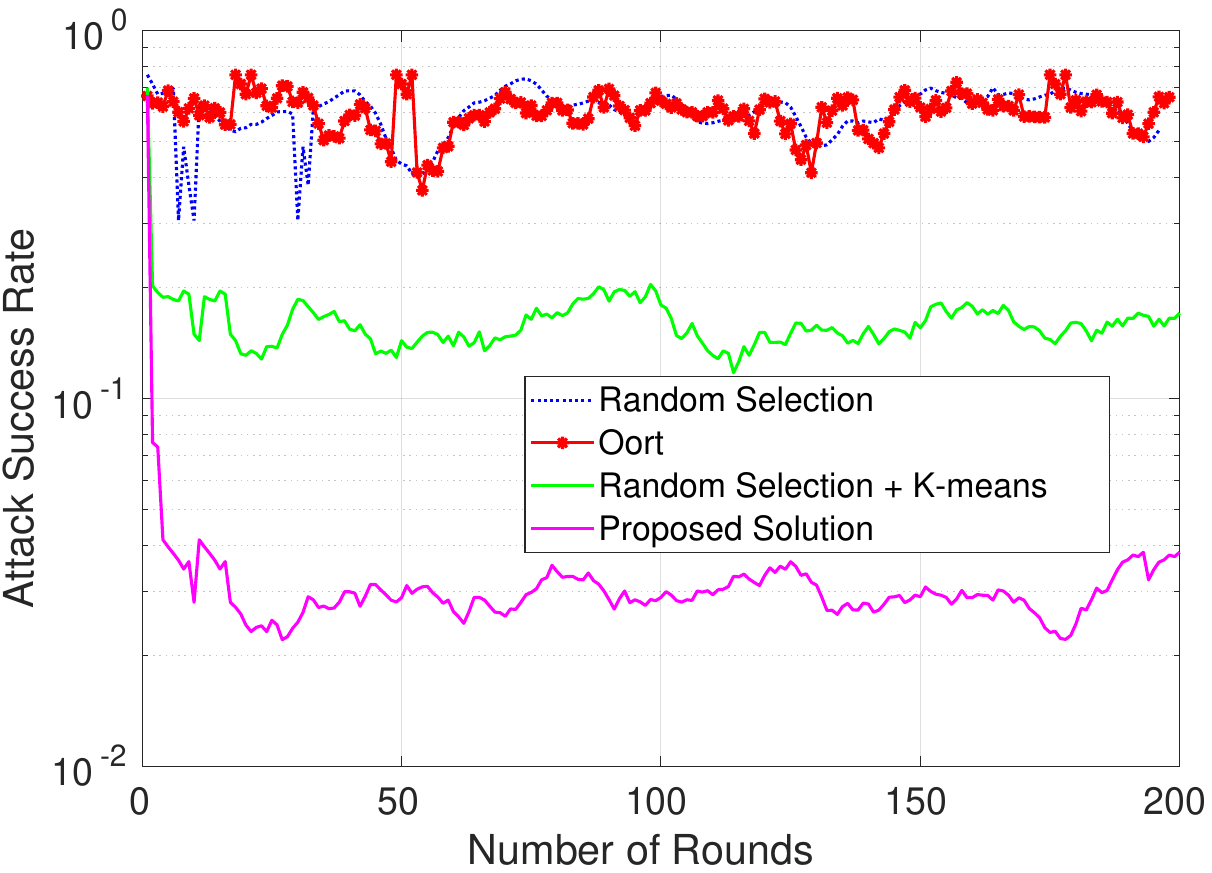}}

    \caption{Attack success rate vs. FL rounds under targeted attack (MNIST dataset).}
    \label{ASRT}
  \end{minipage}
\end{figure}

\begin{figure}[!h]
  \centering

  \begin{minipage}{.85\linewidth}
  
    \centering
    \subcaptionbox{$f$=10\%}
      {\includegraphics[trim={1cm 6.5cm 2.25cm 7.2cm},clip,width=\linewidth]{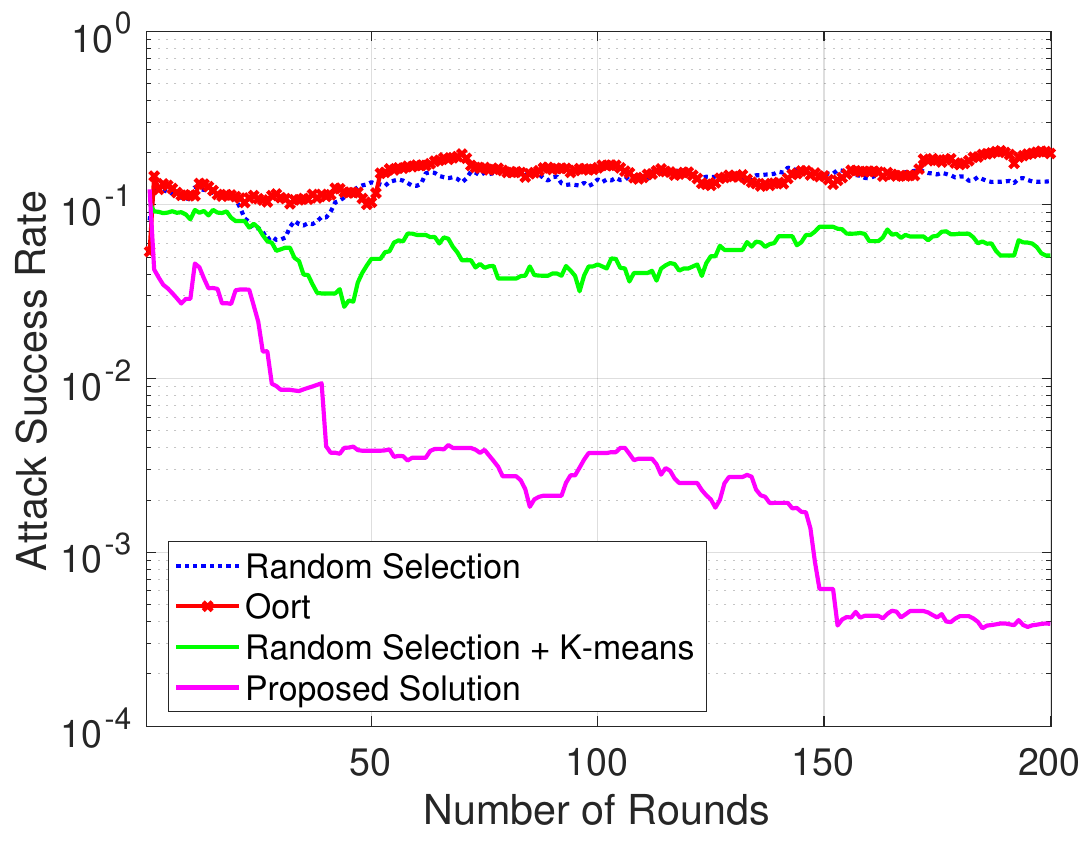}}

    \subcaptionbox{$f$=20\%}
      {\includegraphics[trim={1cm 6.5cm 2.25cm 7.2cm},clip,width=\linewidth]{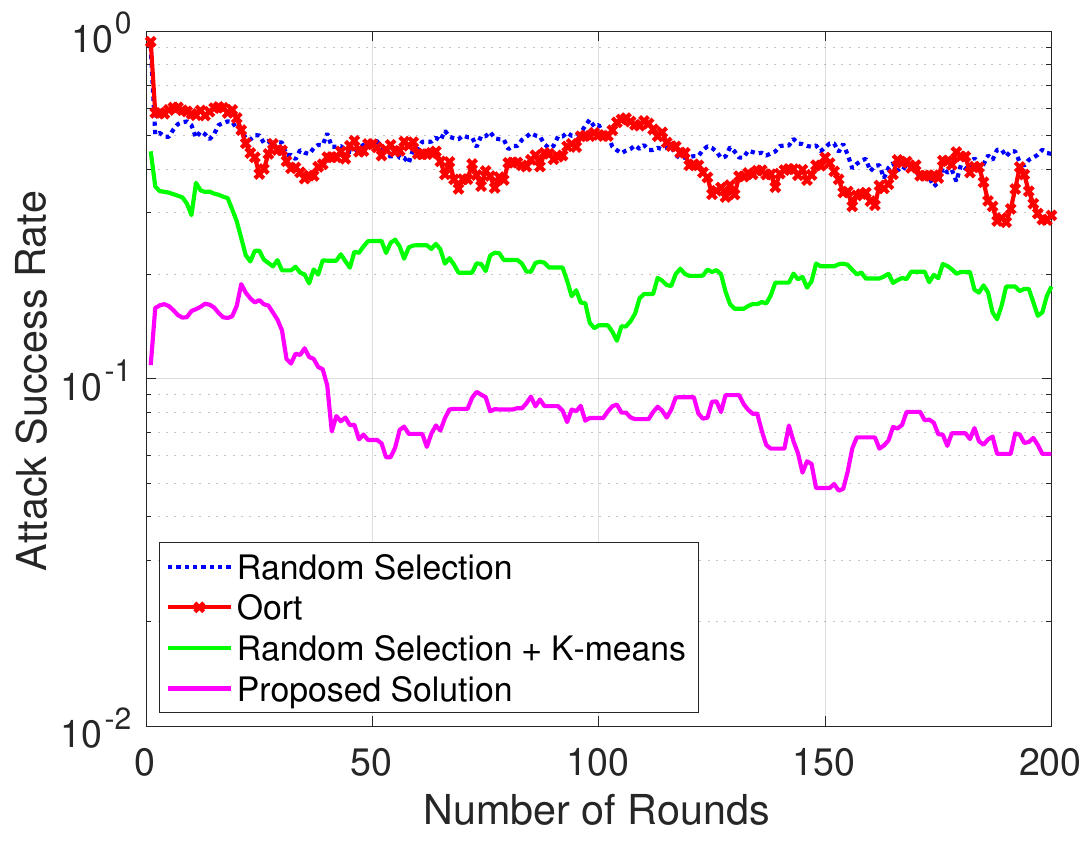}}

    \subcaptionbox{$f$=33\%}
      {\includegraphics[trim={0.25cm 5.5cm 1.25cm 6.3cm},clip,width=\linewidth, height=170pt]{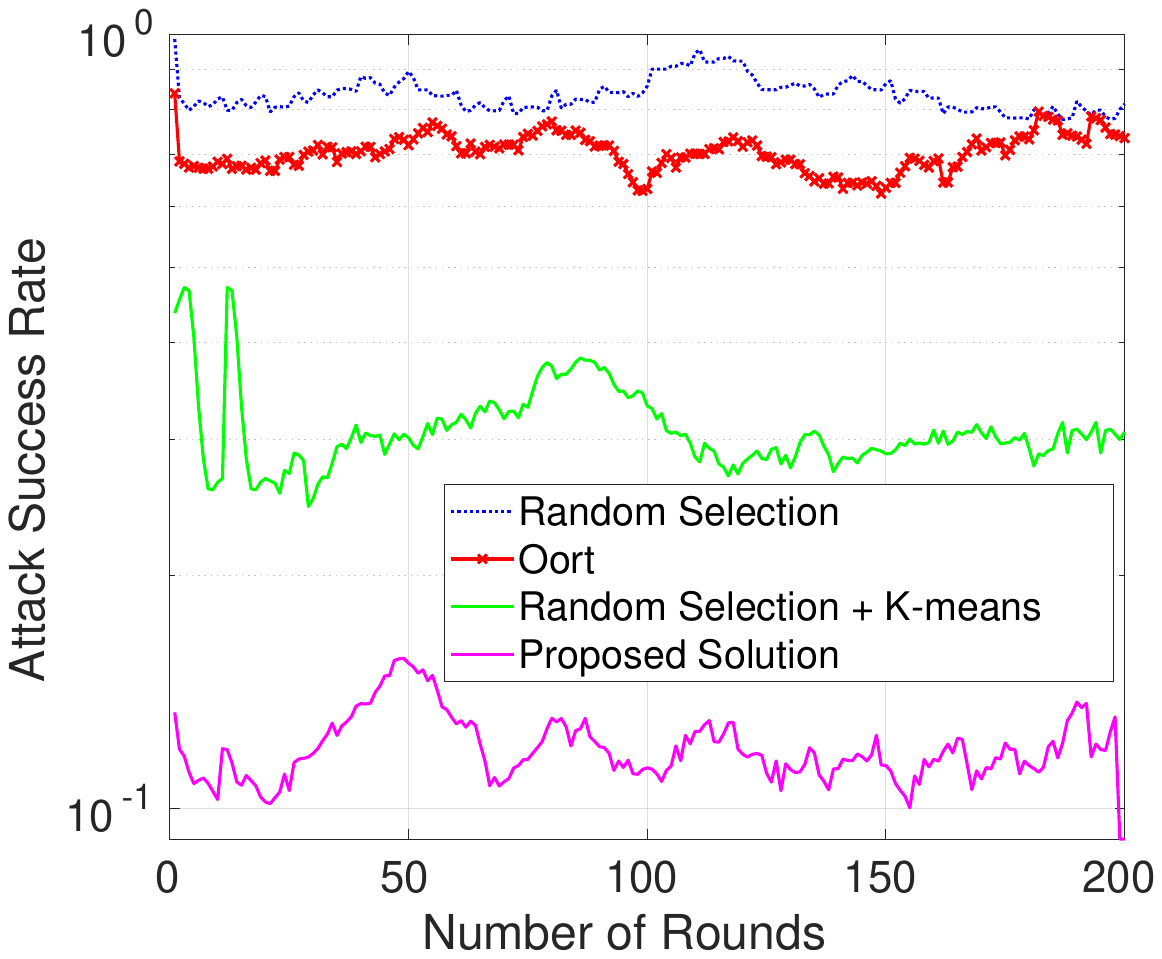}}

    \caption{Attack success rate vs. FL rounds under targeted attack (CIFAR-10 dataset).}
    \label{ASRT-cifar}
  \end{minipage}
\end{figure}

\noindent
\textbf{\textit{Performance under targeted attacks:}}
Under a targeted attack such as flipping labels of a target class, the accuracy for that specific class is affected, especially when $f$ is high. In Figs. \ref{ASRT} and \ref{ASRT-cifar}, we illustrate the
performance, in terms of ASR$_{ta}$, of the different participant selection methods, for the MNIST and CIFAR-10-related datasets, respectively. 
For the MNIST dataset, our method demonstrates a stable (i.e., robust) behavior for any $f$ value at performance around 0.2\%. Meanwhile, the benchmarks' performances are above 2\%. For the CIFAR-10 dataset, the proposed algorithm achieves ASR$_{ta}$ below 1\% for $f\leq 10\%$ and around 7\% for $f>10\%$. Meanwhile, the other approaches realize ASR$_{ta}$ above 10\% for any $f$.   

Unlike the benchmarks, the proposed solution is able to realize correct UAV clustering and efficient malicious participants elimination, even for high $f$ and complex datasets such as the CIFAR-10.
On the other hand, ``Random Selection + K-means'' demonstrated better performances than the other benchmarks but less than the proposed method and with instability through rounds. This is due to the randomness in K-means initialization that affects FP and eliminates non-malicious participants, as shown in Tables \ref{tab:2} and \ref{tab:cifar}. 
Finally, ``Oort'' demonstrates the worst results, similar to ``Random Selection'', since it does not effectively eliminate malicious UAVs. 

In Tables \ref{tab:2} and \ref{tab:cifar}, we compare the results of the above solutions in terms of FN, FP, and accuracy, under different attack scenarios, where targeted and untargeted attacks refer to the events where the ratio of UAVs $f$ is totally dedicated to that specific attack, while the targeted+untargeted scenario means that only $f/2$ are dedicated for untargeted attacks and the other $f/2$ to targeted ones. To be noted that the accuracy refers to the global accuracy when no or only untargeted attacks are performed, while in scenarios of targeted attack and targeted+untargeted attack, two accuracy values are presented for each $f$ value, where the first (top) refers to the global accuracy, while the second (bottom) to the accuracy of the targeted class with the label flipping attack only, i.e., the classes of the number ``5'' and ``cat'' in MNIST and CIFAR-10 experiments, respectively.
Accordingly, the use of DBSCAN in our solution is more efficient in identifying clusters of non-malicious and malicious participants than ``Random selection + K-means'' and ``Oort'' in any scenario. Indeed, due to the K-means' sensitivity to outliers and noise, it tends to provide high FP and FN ratios, in particular at high $f$ values, and thus the system's accuracy is impacted. In contrast, DBSCAN demonstrates stable behavior regardless of the type of attackers, since it is a density-based algorithm. ``Oort'' provides the worst results as it is not using an efficient mechanism to mitigate attacks. 
The impact of both untargeted and targeted attacks have a small impact on the global accuracy of the proposed solution, while ``Random Selection + K-means'' performance significantly degrades with $f$, compared to the no attack scenario. Indeed, this method is inefficient when no attackers are present, thus underperforming, i.e., it is mis-clustering the UAVs. However, when $f>0$, K-means operates well and thus is able to achieve better results. In contrast, ``Oort'' performs very well when no attacks occur, but its accuracy drastically drops as soon as $f>0$. This is mainly due to its incapability to correctly assess the maliciousness of participants.

\begin{table}[!h]
\caption{ Comparison of participant selection methods in terms of FN, FP, and global/class accuracy (MNIST dataset).}
    \centering
\begin{tabular}{|c|c|c|c|c|c|}
\hline
\makecell{\textbf{Attack}\\ \textbf{Type}} & \makecell{$f$ \\ \textbf{(\%)}}
  & \textbf{Selection Method} & \makecell{\textbf{FN}\\ \textbf{(\%)}} & \makecell{\textbf{FP}\\ \textbf{(\%)}} & \makecell{\textbf{Acc.} \\\textbf{(\%)}} \\ 
\hline
\multirow{3}{*}{No attack}    & \multirow{3}{*}{0} 
  & Rand. Sel. + K-means& 0 &  66 & 65.65 \\ \cline{3-6}
&  & Oort & 0 & 98 & 96.03 \\ \cline{3-6}
&  & \textbf{Proposed solution} & \textbf{0}& \textbf{2} & \textbf{97.02}\\
\cline{1-6}

\multirow{8}{*}{\makecell{Untargeted\\ Attack}}    & \multirow{3}{*}{10} 
  & Rand. Sel. + K-means& 0 &  30 & 77.65\\ \cline{3-6}
&  & Oort & 92 & 96 & 10.01\\ \cline{3-6}
&  & \textbf{Proposed solution} & \textbf{0}& \textbf{0.5} & \textbf{97.01}\\ 
\cline{2-6}
&  \multirow{3}{*}{20}& Rand. Sel. + K-means & 50& 10 & 55.32\\ \cline{3-6}
&  & Oort & 88 & 97 & 9.74\\ \cline{3-6}
& & \textbf{Proposed solution} & \textbf{0}& \textbf{0.1} & \textbf{97}\\
\cline{2-6}

&  \multirow{3}{*}{33}& Rand. Sel. + K-means & 70& 33 & 32.82 \\ \cline{3-6}
&  & Oort & 90 & 95 & 5.66\\ \cline{3-6}
&  & \textbf{Proposed solution} & \textbf{0}& \textbf{3} & \textbf{96.8}\\
\cline{2-6}
\hline
\multirow{8}{*}{\makecell{Targeted\\ Attack}}    & \multirow{3}{*}{10} 
  & Rand. Sel. + K-means& 0 & 50 & \makecell{68.67 \\ 92.15}  \\ \cline{3-6}
&  & Oort & 91 & 94 & \makecell{10.78 \\ 85.02} \\ \cline{3-6}
&  & \textbf{Proposed solution} & \textbf{0}& \textbf{0.1} & \makecell{\textbf{97} \\ \textbf{98.27}} \\
\cline{2-6}
&  \multirow{3}{*}{20}& Rand. Sel. + K-means & 11 & 33 & \makecell{60.01 \\ 85.75} \\ \cline{3-6}
&  & Oort & 90 & 92 & \makecell{9.35 \\ 64.22} \\ \cline{3-6}
& & \textbf{Proposed solution} & \textbf{0.2}& \textbf{0.5} & \makecell{\textbf{96.97} \\ \textbf{98.22}}\\
\cline{2-6}

&  \multirow{3}{*}{33}& Rand. Sel. + K-means & 37 & 30 & \makecell{55.37 \\ 83.56}  \\ \cline{3-6}
&  & Oort & 96& 98 & \makecell{6.78 \\ 45.85} \\ \cline{3-6}
&  & \textbf{Proposed solution} & \textbf{0.2}& \textbf{1} & \makecell{\textbf{96.86} \\ \textbf{98.19}}\\
\cline{2-6}
\hline

\multirow{8}{*}{\makecell{Targeted \\ + Untargeted \\ Attack}}    & \multirow{3}{*}{10} 
  & Rand. Sel. + K-means& 0 &  40 & \makecell{66.20 \\ 94.25} \\ \cline{3-6}
&  & Oort & 90 & 95 & \makecell{9.22 \\ 66.80} \\ \cline{3-6}
&  & \textbf{Proposed solution} & \textbf{0} & \textbf{0.1} & \makecell{\textbf{97} \\ \textbf{98.29}}\\
\cline{2-6}
&  \multirow{3}{*}{20}& Rand. Sel. + K-means & 30& 10& \makecell{58.90 \\ 89.89}\\ \cline{3-6}
&  & Oort & 95 & 93& \makecell{8.07 \\ 66.23}\\ \cline{3-6}
& & \textbf{Proposed solution} & \textbf{0.1} & \textbf{0.4} & \makecell{\textbf{96.98} \\ \textbf{98.25}}\\
\cline{2-6}

&  \multirow{3}{*}{33}& Rand. Sel. + K-means & 50 & 30& \makecell{53.89 \\ 90.24}\\ \cline{3-6}
&  & Oort & 98& 95& \makecell{5.87 \\ 50.20}\\ \cline{3-6}
&  & \textbf{Proposed solution} & \textbf{0.5}& \textbf{2}& \makecell{\textbf{96.84} \\ \textbf{98.21}}\\
\cline{2-6}
\hline
\end{tabular}
\label{tab:2}
\end{table}

\begin{table}[H]
\caption{ Comparison of participant selection methods in terms of FN, FP, and global/class accuracy (CIFAR-10 dataset).}
    \centering
\begin{tabular}{|c|c|c|c|c|c|}
\hline
\makecell{\textbf{Attack}\\ \textbf{Type}} & \makecell{$f$ \\ \textbf{(\%)}}
  & \textbf{Selection Method} & \makecell{\textbf{FN}\\ \textbf{(\%)}} & \makecell{\textbf{FP}\\ \textbf{(\%)}} & \makecell{\textbf{Acc.} \\\textbf{(\%)}} \\ 
\hline
\multirow{3}{*}{No attack}    & \multirow{3}{*}{0} 
  & Rand. Sel. + K-means& 0 &  71 & 45.03 \\ \cline{3-6}
&  & Oort & 0 & 97.3 & 55.17 \\ \cline{3-6}
&  & \textbf{Proposed solution} & \textbf{0}& \textbf{5} & \textbf{57.5}\\
\cline{1-6}

\multirow{8}{*}{\makecell{Untargeted\\ Attack}}    & \multirow{3}{*}{10} 
  & Rand. Sel. + K-means& 45 &  33 & 47.60\\ \cline{3-6}
&  & Oort & 93 & 95 & 8.01\\ \cline{3-6}
&  & \textbf{Proposed solution} & \textbf{0.2}& \textbf{0.1} & \textbf{56.82}\\ 
\cline{2-6}
&  \multirow{3}{*}{20}& Rand. Sel. + K-means & 48 & 36 & 42.32\\ \cline{3-6}
&  & Oort & 85 & 98 & 6.23\\ \cline{3-6}
& & \textbf{Proposed solution} & \textbf{0.5}& \textbf{1} & \textbf{54.90}\\
\cline{2-6}

&  \multirow{3}{*}{33}& Rand. Sel. + K-means & 70& 33 & 39.82 \\ \cline{3-6}
&  & Oort & 91 & 96 & 5.22\\ \cline{3-6}
&  & \textbf{Proposed solution} & \textbf{2}& \textbf{3} & \textbf{54.51}\\
\cline{2-6}
\hline
\multirow{8}{*}{\makecell{Targeted\\ Attack}}    & \multirow{3}{*}{10} 
  & Rand. Sel. + K-means& 3 &  48 & \makecell{45.67 \\ 89.02}  \\ \cline{3-6}
&  & Oort & 90  & 95  & \makecell{9.57 \\ 57.3} \\ \cline{3-6}
&  & \textbf{Proposed solution} & \textbf{0.1}& \textbf{0.3} &  \makecell{\textbf{55.98} \\ \textbf{90.59}}   \\
\cline{2-6}
&  \multirow{3}{*}{20}& Rand. Sel. + K-means & 33 &  10 & \makecell{43.53 \\ 86.5}  \\ \cline{3-6}
&  & Oort &  93 & 96  &  \makecell{7.75 \\ 53.65} \\ \cline{3-6}
& & \textbf{Proposed solution} & \textbf{2}& \textbf{ 1.1} & \makecell{\textbf{55.45} \\ \textbf{90.32}} \\
\cline{2-6}

&  \multirow{3}{*}{33}& Rand. Sel. + K-means & 52 & 22  & \makecell{40.65 \\ 83.62} \\ \cline{3-6}
&  & Oort & 95 & 98 & \makecell{6.56 \\ 45.22} \\ \cline{3-6}
&  & \textbf{Proposed solution} & \textbf{3}& \textbf{ 2} & \makecell{\textbf{54.89} \\ \textbf{90.21}}  \\
\cline{2-6}
\hline
\multirow{8}{*}{\makecell{Targeted \\ + Untargeted \\ Attack}}    & \multirow{3}{*}{10} 
  & Rand. Sel. + K-means&  2 &  43  & \makecell{45.01 \\ 88.02}\\ 
  \cline{3-6}
&  & Oort & 89 & 97 & \makecell{6.33 \\ 53.63} \\ \cline{3-6}
&  & \textbf{Proposed solution} & \textbf{0.1}& \textbf{0.2}& \makecell{\textbf{55.80} \\ \textbf{91.32}}\\
\cline{2-6}
&  \multirow{3}{*}{20}& Rand. Sel. + K-means & 39 & 20 & \makecell{43.20 \\ 87.89}\\ \cline{3-6}
&  & Oort & 95 & 93& \makecell{5.88 \\ 51.20}\\ \cline{3-6}
& & \textbf{Proposed solution} & \textbf{0.4}& \textbf{1.3}&\makecell{\textbf{55.22} \\ \textbf{90.75}}\\
\cline{2-6}

&  \multirow{3}{*}{33}& Rand. Sel. + K-means & 65 & 22& \makecell{40.01 \\ 85.24}\\ \cline{3-6}
&  & Oort & 99& 92& \makecell{5.55 \\ 48.51}\\ \cline{3-6}
&  & \textbf{Proposed solution} & \textbf{2} & \textbf{3} & \makecell{\textbf{54.65} \\ \textbf{90.33}}\\
\cline{2-6}
\hline
\end{tabular}
\label{tab:cifar}
\end{table}

\begin{figure}[!h]
    
    \begin{minipage}{\linewidth}
    \centering
   
\includegraphics[trim={0.1cm 0.3cm 0.5cm 0.5cm},clip,width=0.95\linewidth]{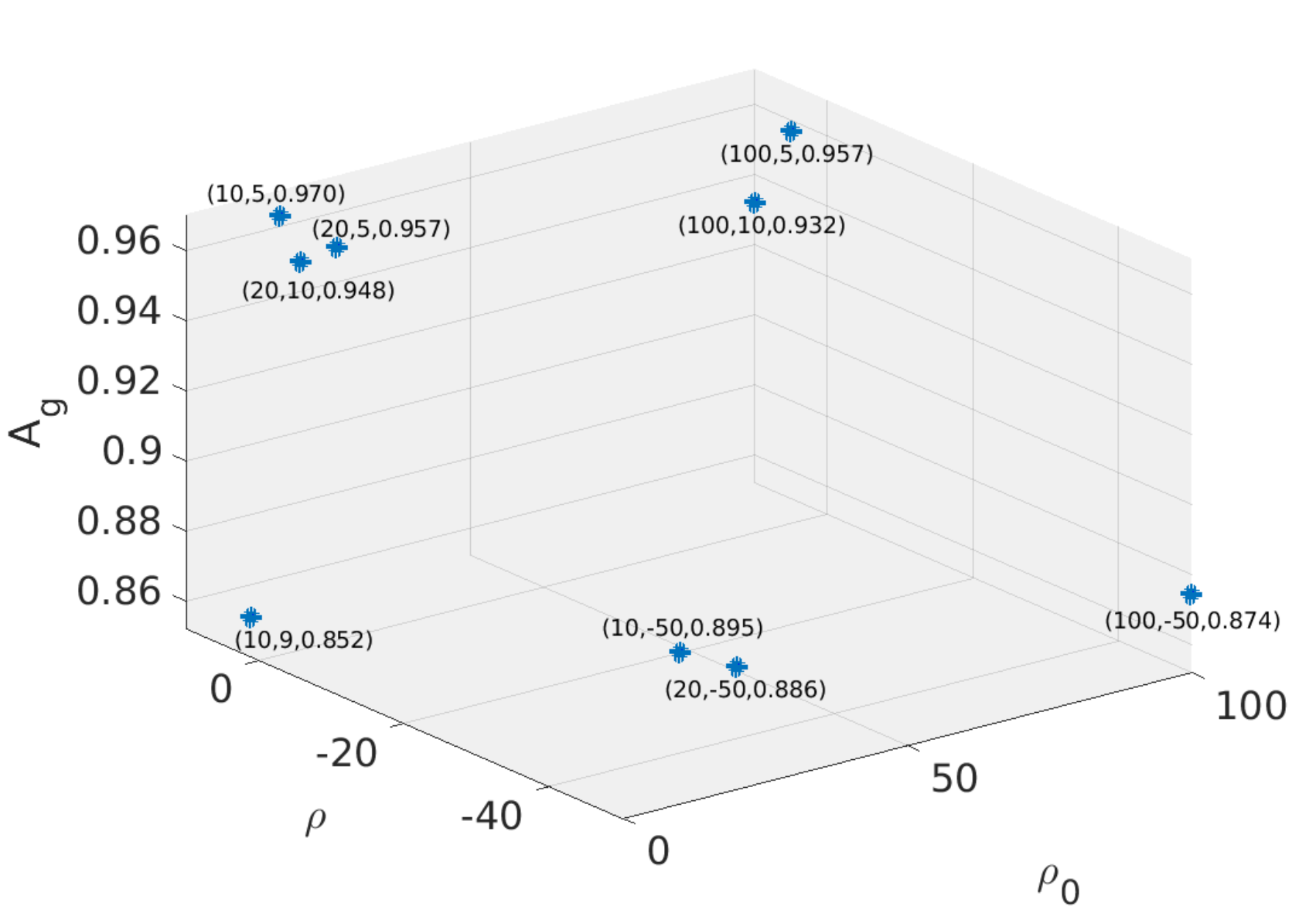}

\caption{Global accuracy vs. $\rho$ and $\rho_0$ (MNIST dataset, Distribution 1).}
\label{rho}
\end{minipage}
\end{figure}

\subsection{Discussion on Hyperparameters Sensitivity}

To enforce a criterion for the minimal contribution from the participants, we established the reputation threshold $\rho$ and its maximal value $\rho_0$ as settable coefficients. Participants having reputations below $\rho$ are not considered in subsequent FL rounds. In order to understand the impact of the values of $\rho$ and $\rho_0$, we illustrate in Fig. \ref{rho} the achieved global accuracy performance as a function of $\rho$ and $\rho_0$, and under the distribution 1 setting. We notice that the choice of $\rho$ and $\rho_0$ values have a significant impact on the global accuracy $A_g$. For instance, a small $\rho$ degrades $A_g$ for any $\rho_0$, which is the result of a very slow process of unreliable UAVs elimination. Another negative impact on $A_g$ is the choice of close $\rho$ and $\rho_0$ values, e.g., $\rho=9$ and $\rho_0=10$ respectively. In this case, several UAVs will be eliminated, seen as with very low reliability scores or falsely being identified as dropouts. In contrast, a large $\rho_0$ value may lead to clients' over-selection. For instance, if a UAV begins as a reliable client before turning unreliable after a certain number of rounds, it will still have a high reliability score, which will have a detrimental effect on the accuracy performance. Therefore, a tradeoff between $\rho$ and $\rho_0$ is necessary to improve $A_g$.

\section{Conclusion}
\label{sec: Conclusion}
In this paper, we have investigated the problem of participant selection in FL under the presence of misbehaving UAVs, by considering straggler, dropout, and malicious participants' profiles. To reach efficient and reliable FL, we proposed a novel three-step selection method that sequentially eliminates misbehaving participants, starting with stragglers, then dropouts, and finally malicious ones. 
Through experiments, we demonstrate the superiority of our solution in terms of accuracy, convergence, average round time, and attack success rate, compared to baseline methods, and considering different data distribution among UAVs and under the absence or presence of malicious nodes.
As a future work, we will extend our investigation on FL to other types of attacks and heterogeneous systems, where participants are of different types and operate under varying constraints. 

\section*{ACKNOWLEDGMENT}
This research is supported by the International Study in Canada Scholarship (SiCS).
\bibliographystyle{IEEEtran}
\bibliography{bibfile}

\begin{IEEEbiography}[{\includegraphics[width=1in,height=1.25in,clip,keepaspectratio]{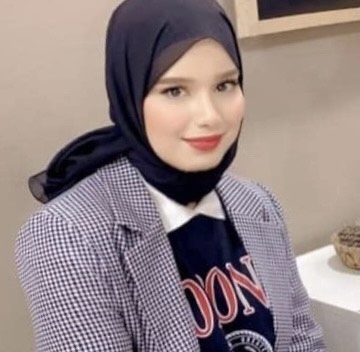}}]{Youssra Cheriguene } received her
M.Sc degree in Computer Science in 2020 at the University of
Laghouat, Algeria. She is currently a Ph.D. student at the
Informatics and Mathematics
Laboratory (LIM) at the
University of Laghouat. Her
research focuses Mobile Edge
Computing, Unmanned Aerial
Vehicles,
and
Federated
Learning. 
\end{IEEEbiography}

\begin{IEEEbiography}[{\includegraphics[width=1in,height=1.25in,clip,keepaspectratio]{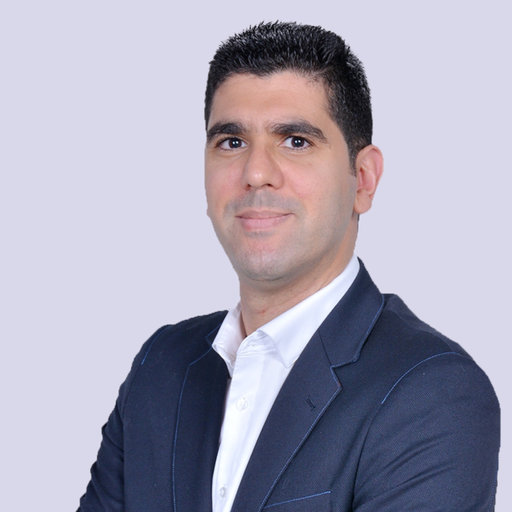}}]{Wael Jaafar }  (S’08, M’14, SM’20) is an Associate Professor at the Software and IT Engineering Department of École de Technologie Supérieure (ÉTS), University of Quebec, Montreal, Canada since September 2022. He holds Master and PhD degrees from Polytechnique Montreal, Canada. Between 2019 and 2022, Dr. Jaafar was with the Systems and Computer Engineering Department of Carleton University as an NSERC Postdoctoral Fellow. From 2014 to 2018, he has pursued a career in the telecommunications industry, where he has been involved in designing telecom solutions for projects across Canada and abroad.  
During his career, Dr. Jaafar was a visiting researcher at Khalifa University, Abu Dhabi, UAE in 2019, Keio University, Japan in 2013, and UQAM, Canada in 2007. He is the recipient of several prestigious grants including NSERC Alexandre-Graham Bell scholarship, FRQNT internship scholarship, and best paper awards at IEEE ICC 2021 and ISCC 2023. His current research interests include wireless communications, integrated terrestrial and non-terrestrial networks, resource allocation, edge caching and computing, and machine learning for communications and networks.
\end{IEEEbiography}

\begin{IEEEbiography}[{\includegraphics[width=1in,height=1.25in,clip,keepaspectratio]{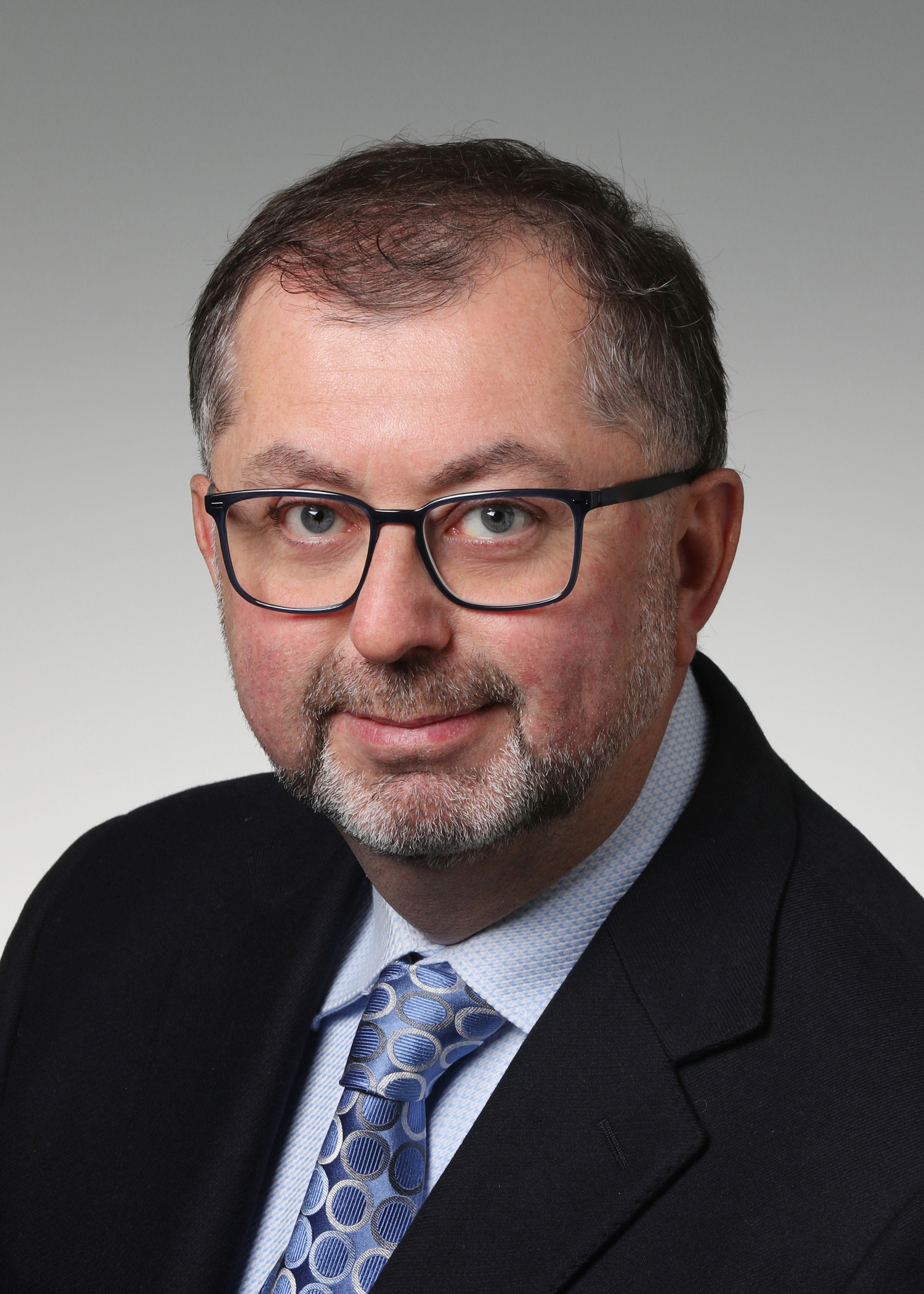}}]{Halim Yanikomeroglu } (Fellow, IEEE)
received the B.Sc. degree in electrical and elec-
tronics engineering from Middle East Technical
University, Ankara, Turkey, in 1990, and the
M.A.Sc. degree in electrical engineering and the
Ph.D. degree in electrical and computer engineer-
ing from the University of Toronto, Canada, in
1992 and 1998, respectively.
Since 1998, he has been with the Department
of Systems and Computer Engineering, Carleton
University, Ottawa, Canada, where he is currently a Chancellor’s Professor.
He has given more than 110 invited seminars, keynotes, panel talks, and
tutorials in the last five years. He has supervised or hosted more than
150 postgraduate researchers in the laboratory with Carleton University. His
extensive collaborative research with industry resulted in 39 granted patents.
His research interests include wireless communications and networks, with
a special emphasis on non-terrestrial networks (NTN) in recent years.
Dr. Yanikomeroglu is a fellow of the Engineering Institute of Canada
(EIC) and the Canadian Academy of Engineering (CAE). He is also a
member of the IEEE ComSoc Governance Council, IEEE ComSoc GIMS,
IEEE ComSoc Conference Council, and IEEE PIMRC Steering Committee.
He received several awards for research, teaching, and service, including the
IEEE ComSoc Satellite and Space Communications TC Recognition Award,
in 2023, the IEEE ComSoc Fred W. Ellersick Prize, in 2021, the IEEE VTS
Stuart Meyer Memorial Award, in 2020, and the IEEE ComSoc Wireless
Communications TC Recognition Award, in 2018. He also received the Best
Paper Award from the IEEE Competition on Non-Terrestrial Networks for
B5G and 6G, in 2022 (grand prize), IEEE ICC 2021, and IEEE WISEE
2021 and 2022. He served as the general chair and the technical program
chair for several IEEE conferences. He is also serving as the Chair for the
Steering Committee of IEEE’s Flagship Wireless Event and the Wireless
Communications and Networking Conference (WCNC). He has also served
on the editorial boards for various IEEE periodicals. He is a Distinguished
Speaker of the IEEE Communications Society and the IEEE Vehicular
Technology Society, and an Expert Panelist of the Council of Canadian
Academies (CCA|CAC). He has also served on editorial boards of various
IEEE periodicals.
\end{IEEEbiography}

\begin{IEEEbiography}[{\includegraphics[width=1in,height=1.25in,clip,keepaspectratio]{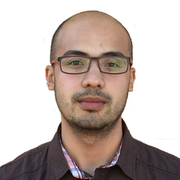}}]{Chaker Abdelaziz Kerrache }
is an Associate Professor at the Department of Computer Science and the head of the Informatics and Mathematics Laboratory (LIM) at the University of Laghouat, Algeria. He received his MSc. degree in Computer Science degree at the University of Laghouat, Algeria, in 2012, and his Ph.D. degree in Computer Science degree at the University of Laghouat, Algeria, in 2017.  His research activity is related to Trust and Risk Management, Secure Multi-hop Communications, Vehicular Networks, Named Data Networking (NDN), and UAVs.
\end{IEEEbiography}

\end{document}